\documentclass{article}
\usepackage{default}

\usepackage[small]{caption}
\usepackage{graphicx}
\usepackage{amsmath}
\usepackage{subcaption}

\usepackage{algorithm}
\usepackage{algorithmic}
\usepackage[T1]{fontenc}

\title{Neural Network Retraining for Model Serving}

\author{
Diego Klabjan
\and
Xiaofeng Zhu\and
\affiliations
Northwestern University, Evanston, IL\\
\emails
d-klabjan@northwestern.edu,
xiaofengzhu2013@u.northwestern.edu
}

\begin{document}
\maketitle

\begin{abstract}
We propose incremental (re)training of a neural network model to cope with a continuous flow of new data in inference during model serving. As such, this is a life-long learning process. We address two challenges of life-long retraining: catastrophic forgetting and efficient retraining. If we combine all past and new data it can easily become intractable to retrain the neural network model. On the other hand, if the model is retrained using only new data, it can easily suffer catastrophic forgetting and thus it is paramount to strike the right balance. Moreover, if we retrain all weights of the model every time new data is collected, retraining tends to require too many computing resources. To solve these two issues, we propose a novel retraining model that can select important samples and important weights utilizing multi-armed bandits. To further address forgetting, we propose a new regularization term focusing on synapse and neuron importance. We analyze multiple datasets to document the outcome of the proposed retraining methods. Various experiments demonstrate that our retraining methodologies mitigate the catastrophic forgetting problem while boosting model performance.
\end{abstract}

\section{Introduction}
\label{4-Introduction}
Powered by deep learning, artificial intelligence is exceeding human intelligence in several tasks. There are still challenges, as training a deep neural network requires substantial data, computing resources, and it does not generalize well. Model training and serving is not a one-time task but an incremental learning process. Once an initial model is well-trained on historical data, it is then periodically fine-tuned or retrained based on a continuous flow of new data for inference in model serving. New data may be collected every second, day, or week. In model serving, there are two important decisions: when to retrain the model and how to efficiently retrain it. We focus on the latter aspect. Retraining a model using only new data can lead to catastrophic forgetting \cite{french1999catastrophic,kirkpatrick2017overcoming}, i.e., the model forgets the knowledge acquired in the past. It is a common practice that a model is retrained on a periodic basis using all old data (data used the last time the model has been (re)trained) and new data (data acquired since the last time the model has been (re)trained). However, this strategy becomes infeasible as data accumulates during model serving. Our study focuses on efficiently retraining a trained neural network model with new data. The amount of old data a retraining process can access is selected dynamically and is subject to computation efficiency.

We study how to efficiently retrain a model from three aspects: mitigating catastrophic forgetting by identifying important neurons, strategically buffering data, and dynamically re-optimizing weights. We assume the following setting to address these aspects. A model is initially trained with some training data and then fine-tuned continually based on (a small amount of) new data. Fine-tuning is triggered periodically, and its timing is not the scope of this work. Every fine-tuning of a model is a retraining session. Old and new data are relative to the incumbent retraining session. 

Catastrophic forgetting is a major barrier for deep neural networks to learn continually. There have been many attempts to limit forgetting. Some existing studies focus on consolidating synapses that are important to the trained model. Weight importance can be measured using the diagonal of the Fisher matrix \cite{kirkpatrick2017overcoming} or gradient magnitudes of weights \cite{aljundi2018memory}. If weights are required to be stable during retraining by imposing regularization, it can prevent the model from learning new patterns in new data. Moreover, both neurons and weights affect model outcome. We introduce a new regularization term to encourage weight updates as long as the neurons do not incur dramatic changes. Inspired by the discussion of representation sparsity by  \cite{aljundi2018selfless}, we present a more efficient regularization term that captures both the importance of neurons and synapses/weights. The other line of research to cope with forgetting is to dynamically adjust the underlying network architecture \cite{YuSlimmable,Rusu2016}. In the context of model serving, this is inefficient since current AutoML strategies require weeks of training.

As \cite{Mehta2018,lopez2017gradient}, and \cite{kemker2018measuring} conclude, memory replay approaches generally outperform regularization-based approaches. Given a limited data/memory buffer size, memory replay approaches tune a model with new data as well as a small subset of old data. \cite{Mehta2018} compare several competitive methods, e.g., clustering and herding, to select important individual samples from old data. As the authors state, individual sample selection can be computationally expensive and can be easily influenced by outlier data. Moreover, the same training samples in different training epochs can have very different stochastic gradient updates. To this end, we build a reward system based on loss decreases or weight magnitude changes and use a multi-armed bandit (MAB) algorithm to select batches of old data that are influential in loss function optimization. Each arm corresponds to a mini-batch with reward being the loss decrease or the weight magnitude change, and the reward is observed only after an arm is selected (a mini-batch is optimized). We use the standard epoch-based weight optimization to warm up weights then use an MAB algorithm to select mini-batches. Mini-batches selected most often in the current (re)training session are used in the next retraining session. We showcase superior results of the MAB-based sampling method. We also demonstrate that the combination of the MAB-based memory replay method and regularization can boost the effectiveness of retraining.

Meta-learning in terms of directly or indirectly changing network structures to address weight optimization is another major innovation in continual learning. Network compression and weight sharing among different tasks/domains are common ways of reducing the number of trainable parameters. Since continual learning aims to enhance a trained model, prior studies address ensemble ideas of expanding trained networks including AutoML. In addition, we have already argued that AutoML-like methods are too slow. In our retraining method, we do not consider additional trainable parameters as we keep the architecture fixed but introduce a novel way of tuning a subset of weights at a time. We cluster weights after each (re)training session given weight changes in consecutive epochs. In the subsequent retraining session, we use another reward system based on loss decreases or weight magnitude changes where each arm corresponds to a cluster of weights in an MAB algorithm. In each retraining step, we use an MAB algorithm to select an arm/cluster and only optimize the weights in this cluster in a mini-batch. Then, we calculate the reward as the amount of loss decrease or weight magnitude change. To this end, only a small portion of weights receive gradient updates, and therefore, computing resources are allocated dynamically.

The proposed MAB-based retraining methodology with the addressed three components outperforms the models that only rely on regularization terms with reservoir sampling (a state-of-the-art memory replay method) \cite{{vitter1985random}} and standard gradient optimization on average by 0.48\%. The improvements range from 0.07\% to 1.53\% on a variety of network architectures including fully connected, convolutional, and recurrent networks. Data samples selected based on MAB yield a better model performance on average by 0.13\% over reservoir sampling. Strategically optimizing a subset of weights using MAB with clustering  improves model performance by 0.29\% over standard optimization where all weights are optimized for every mini-batch. It turns out that MAB-based optimization produces solutions that offer better generalizations.

The major contribution of our work is the development and integration of the addressed three strategies: mitigating catastrophic forgetting, strategically buffering old data, and dynamically optimizing weights by means of clustering and MAB. The retraining model not only mitigates catastrophic forgetting but also performs well on new data, i.e. it generalizes better. The model is generic and can be applied to any (re)trained neural architecture with any loss function. In summary, we present a simple way of mitigating catastrophic forgetting by regularizing neuron changes, which also boosts model performance on new data; we enhance memory replay using MAB; and we propose a novel way of dynamically optimizing weights using clustering and MAB.

The remainder of this paper is organized as follows. In Section \ref{4-Related Work}, we explain past studies related to our work. In Section \ref{Neural Network Retraining Methodology}, we detail the three components in our methodology: synapse and neuron importance, MAB-based memory replay, and MAB-based weight optimization. In Section \ref{4-Experiments}, we introduce the datasets and experimental settings and demonstrate the results. In the end, we conclude in Section \ref{4-Conclusion}.

\section{Related Work}
\label{4-Related Work}

In this section, we distinguish our study from related research areas: continual learning and multi-task/sequential learning. We also detail competitive techniques that are proposed for solving catastrophic forgetting. Lastly, we introduce popular MAB algorithms used in our models.

\subsection{Continual Learning}

Most continual learning studies, especially multi-task/sequential learning studies, do not allow access to old data \cite{li2017learning,aljundi2018selfless}. Given this restriction, only regularization techniques can be applied. General continual learning focuses on retaining knowledge acquired from old tasks by studying task ordering and parameter shifting. The objective in retraining is to tune a previously (re)trained model to have a good performance on both new and old tasks (data in terms of model serving). General continual learning studies focus on the aspect of changes in tasks and thus the aforementioned prior works rely on the presence of different tasks \cite{swaroop2019improving}, but this is not the goal of our study. For this reason, continual learning studies focusing exclusively on multiple tasks are not applicable to the process of retraining.

\subsection{Catastrophic Forgetting}
\label{Catastrophic Forgetting}
We next introduce three major advances in solving catastrophic forgetting: regularization, memory consolidation, and ensemble networks.

In regularization studies, when the weights of a trained network are being tuned with new data, weights that are important to the previous training session are kept relatively stable to maintain the performance on old data. The original loss function is then combined with a regularization term to penalize updates of the important weights when retraining on a new session or task. Regarding measuring the importance of weights, \cite{kirkpatrick2017overcoming} propose the elastic weight consolidation (EWC) algorithm, an established benchmark model that utilizes the diagonal of the Fisher matrix. In particular, \cite{aljundi2018memory} propose the memory aware synapses (MAS) model, another well-known benchmark model, by measuring how different weights influence model outputs. \cite{aljundi2018selfless} make a breakthrough by introducing sparsity at the neuron level and propose the selfless learning (Selfless) model. In spite of their effectiveness, the EWC and MAS models only consolidate weights, which can prevent a model from learning using new data. The Selfless model takes the relatedness of pairs of neurons into consideration, which is computationally intensive and often prohibitive in model serving. The newly proposed regularization terms in our methodology consider only individual neurons and is thus more computationally efficient. It turns out that the performance is also superior.

Intuitively, if a model can access all old training data in any retraining session, catastrophic forgetting would be maximally reduced. Memory replay studies in continual learning solve catastrophic forgetting by selecting a small portion of old data or by generating synthetic important samples \cite{lopez2017gradient}. These existing memory replay approaches usually require additional training for selecting samples. Reservoir sampling is commonly used for choosing a pre-defined number of random samples without replacement from a population in a single pass \cite{vitter1985random}. It creates a reservoir pool of a fixed size, and it maintains a random uniform distribution when replacing a sample in the pool with a new sample. It is commonly applied in data streaming when it is difficult to fit all samples into memory. Inspired by reinforcement learning, we utilize a more promising sampling approach, MAB, to choose important training batches for network retraining. Sampling is done in each (re)training session in an online learning fashion. Our method of selecting important batches is more practical and efficient than selecting individual samples.

Ensemble networks, sometimes referred to as meta-learning in continual learning, take a different direction to overcome catastrophic forgetting by designing multiple networks for different tasks or expanding trained networks \cite{han2015deep_compression}. The biggest limitation is that memory usage increases with new data or new tasks in training and even inference. Our methodology does not introduce additional parameters, and we use an MAB algorithm to selectively optimize a subset of weights in mini-batch retraining. Unlike models that freeze network layers \cite{brock2017freezeout}, of which some weights are frozen during an entire (re)training session, all weights in our methodology are considered, albeit not in each iteration.

\subsection{Multi-armed Bandits}
\label{Multi-armed Bandits}
We briefly introduce competitive MAB algorithms that are used for creating a reward system in our retraining methodology. The MAB problem is to select an action among a finite number of actions. The reward is observed after the action is executed by the environment. The five commonly used MAB algorithms are expected improvement (EI) \cite{audibert2010best}, upper confidence bound (UCB) \cite{auer2002using}, Thompson sampling (TS) \cite{kaufmann2012thompson}, exponential-weight algorithm for exploration and exploitation (EXP3 and EXP4) \cite{auer2002nonstochastic,beygelzimer2011contextual}, and top-two expected improvement (EI2) \cite{qin2017improving}. We choose the best MAB algorithm in our retraining methodology based on the experimental performance under different settings.

The use of MAB algorithms for continual learning is studied by \cite{graves2017automated}. However, the purpose of their data sampling approach is to overcome forgetting by taking different tasks as arms, which is different from our study where we consider arms as mini-batches or clusters of weights.

\section{Neural Network Retraining Methodology}
\label{Neural Network Retraining Methodology}
In this section, we describe the three components of the new retraining methodology. We let $\theta^m$ denote the model parameters trained on data $D^m$ (the ``old'' data). The newly arrived data is denoted by $D^{m+1}$. After observing $D^{m+1}$, an oracle determines that the model needs to be retrained, and thus the task is to efficiently find model parameters $\theta^{m+1}$ on samples $D^m \bigcup D^{m+1}$ (or an approximately selected subset). Given sample $x$ and ground truth $g$, we let $L_{\theta}(x, g)$ denote the loss function and let $l^{m+1}(\theta)$ denote the objective function of model $m+1$ given parameters $\theta$.

\subsection{Synapse and Neuron Importance}
\label{Synapse and Neuron Importance}
We first lay out the methodology of our regularization term. Given a generic neural network model with $N$ layers $\{Y_j\}_{j=0}^N$, the equations specifying the dynamics are
\begin{equation} \label{neuron importance}
Y_{i+1} = f_{i+1}(W_i Y_i + B_i),
\end{equation}
where $Y_i$ denotes the vector of neurons in layer $i$, $W_i$ and $B_i$ denote the matrix and vector of weights between layer $i$ and layer $i+1$, and $f_{i+1}$ denotes the activation function in layer $i+1$. 
The trainable parameters are $\theta = \{W_i, B_i\}_{i=1}^N$.
We denote neuron values of the trained model by $Y^m_i$ on $D^m$. Note that $Y_i=Y_i(\theta)$, but we explicitly show this dependence only when needed for clarity. A second subscript, when present, relates to individual neurons.

In regularization approaches, in training session $m+1$, weights that are important to training session $m$ are ``consolidated.'' From (\ref{neuron importance}), we can easily conclude that the magnitudes of weights with respect to model outputs are partially influenced by the magnitudes of both neuron activations and weights because of the chain rule

\begin{equation} \label{chain_rule}
\begin{split}
\frac{\partial l^{m+1}}{\partial W_{i-1}} &= \frac{\partial l^{m+1}}{\partial Y_{N}} \frac{\partial Y_{N}}{\partial W_{i-1}} \\
\frac{\partial Y_N}{\partial W_{i-1}} &= \frac{\partial Y_N}{\partial Y_{i}} 
\frac{\partial Y_{i}}{\partial W_{i-1}}. \\
\end{split}
\end{equation}

The proposed loss function is
\begin{equation} \label{neuron consolidation}
\begin{split}
&l^{m+1}(\theta) =  E_{(x,g) \sim p(\cdot |D^m \bigcup D^{m+1})}L_{\theta}(x, g) + \\ 
&\alpha \sum_{i=1}^N \sum_k  \frac{\partial  \left\lVert Y_N\right\rVert_2^2  }{\partial Y_{ik}}\Bigr|_{Y=Y^m} (Y_{ik}(\theta) - Y_{ik}^{m})^2+ \\
&\beta \sum_s  \frac{\partial \left\lVert Y_N(\bar{\theta})\right\rVert_2^2 }{\partial \bar{\theta_s}}\Bigr|_{\bar{\theta}=\theta^m} (\theta_s - \theta_s^{m})^2, \\
\end{split}
\end{equation}
where the first term captures standard loss, the second term captures neuron activities, and the last term regularizes trainable parameters. Values $\alpha$ and $\beta$ are hyperparameters. We weigh the last two terms by gradient values. 
Term $(Y_{ik}(\theta) - Y_{ik}^{m})^2$ captures the change in the neuron activity, which is an approximation to $\frac{\partial Y_{i}}{\partial W_{i-1,k}}$. This term in (\ref{chain_rule}) is multiplied by a linear combination of $\frac{\partial Y_N}{\partial Y_{i}}$ approximated by the L2 norm, which justifies the weights in the regularization terms in (\ref{neuron consolidation}).

Compared to EWC, our regularization terms are based on model outputs, which does not require additional weight importance calculations after each training session. Compared to MAS, we add the second term for regularizing the sensitivity of outputs with respect to neurons. The Selfless model also considers weight and neuron importance, but it has computationally expensive operations of calculating pairwise relatedness of neurons. The weights in (\ref{neuron consolidation}) can be easily computed by backpropagation.

\subsection{MAB-based Memory Replay}
\label{MAB-based Memory Replay}
In this section, we propose a new memory replay method utilizing MAB algorithms. Although regularization methods can marginally solve catastrophic forgetting, we show in our experiments that when integrating with an adequate memory replay method, the retraining performance can be boosted notably. We propose an online memory replay algorithm that selects optimal mini-batches in training session $m+1$ for training session $m+2$.

\label{Optimal Mini-batch Sampling Using Multi-armed Bandits}
\begin{figure*}[pt]
\begin{subfigure}[b]{0.5\textwidth}
  \includegraphics[width = \textwidth]{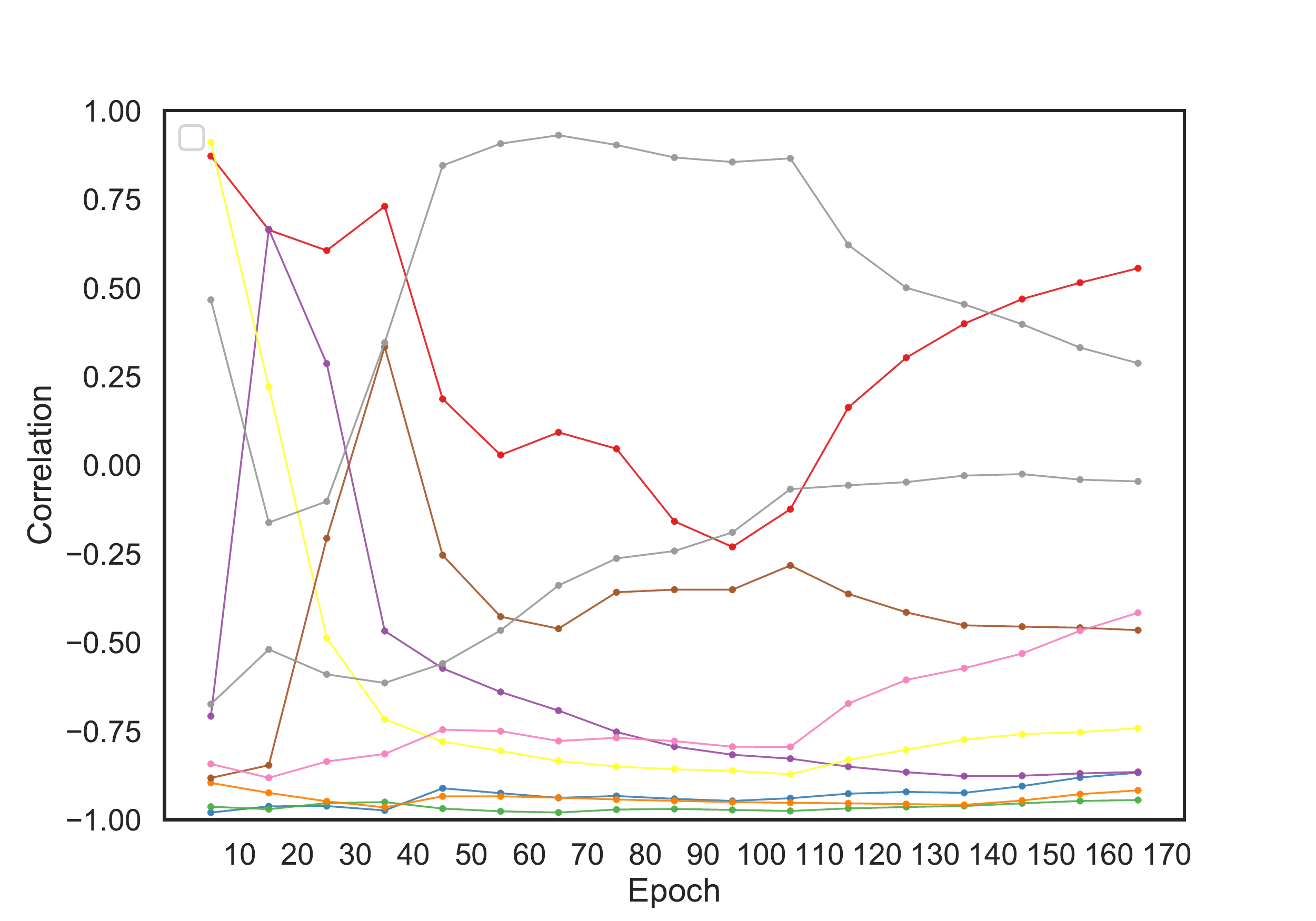}
  \caption{The 1st CNN layer}
  \label{fig:The first layer}
\end{subfigure}%
\begin{subfigure}[b]{0.5\textwidth}
  \includegraphics[width =\textwidth]{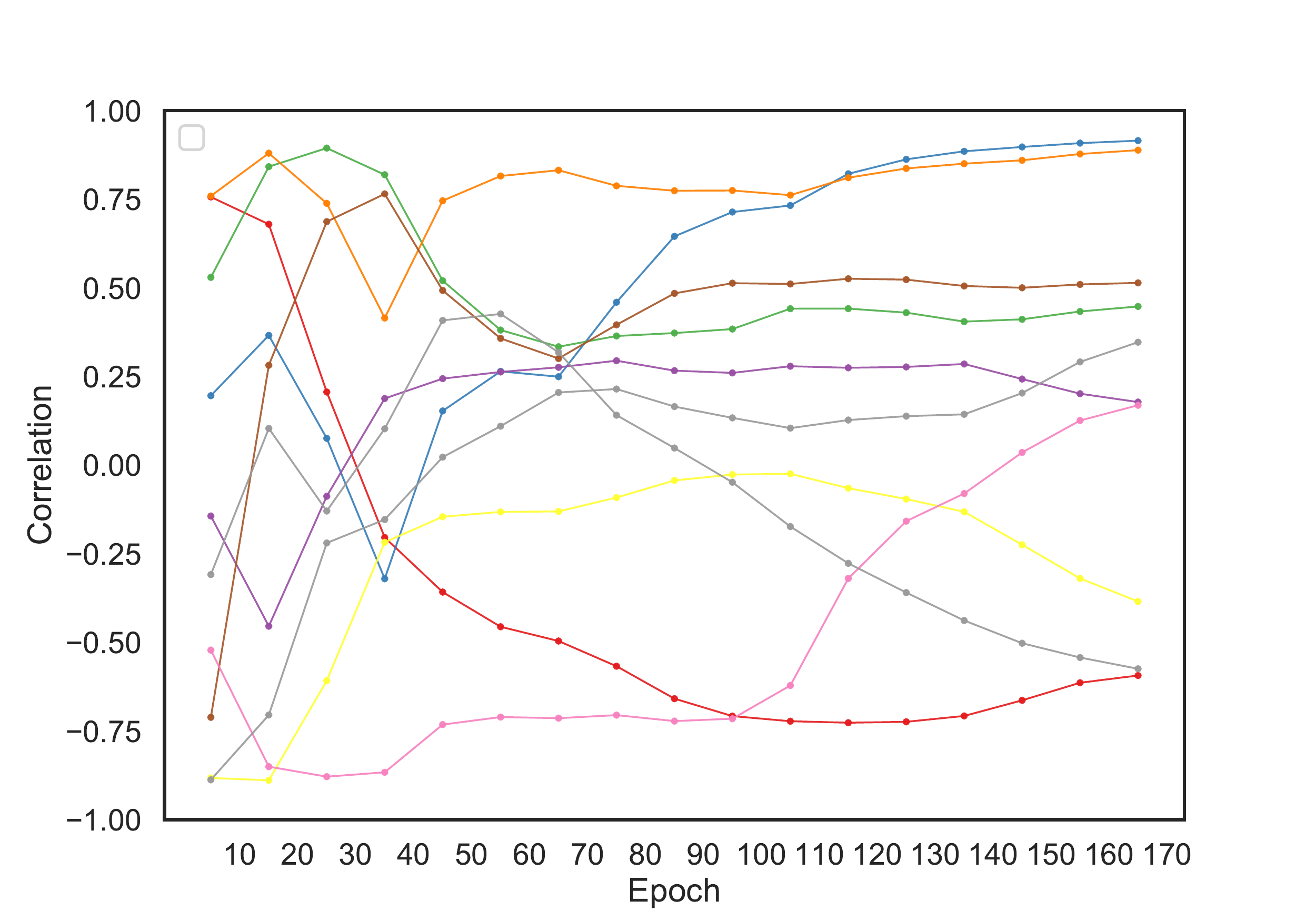}
  \caption{The 2nd CNN layer}
  \label{fig:The second layer}
\end{subfigure}%

\begin{subfigure}[b]{0.5\textwidth}
  \includegraphics[width = \textwidth]{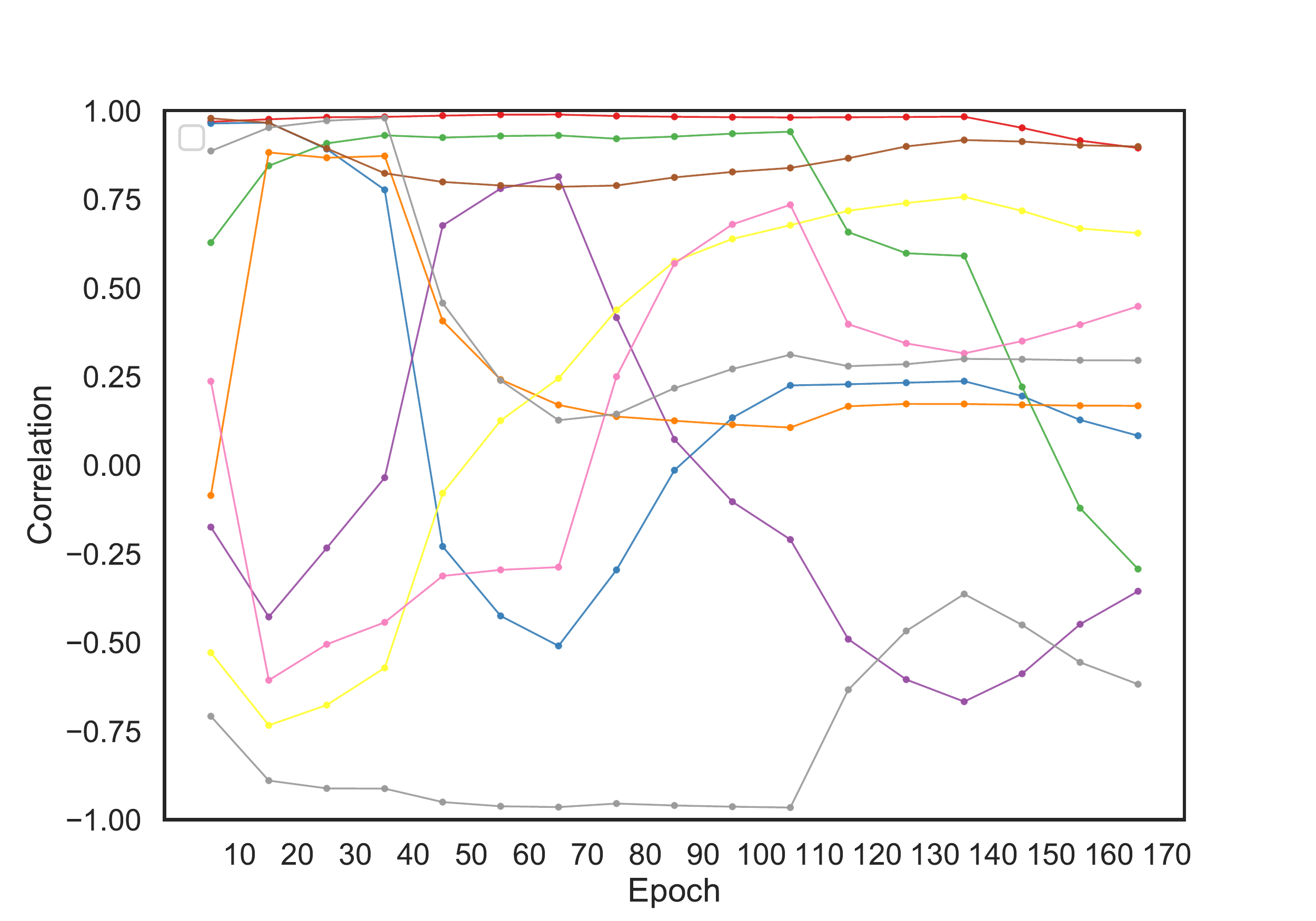}
  \caption{The 1st FC layer}
  \label{fig:The third layer}
\end{subfigure}%
\begin{subfigure}[b]{0.5\textwidth}
  \includegraphics[width =\textwidth]{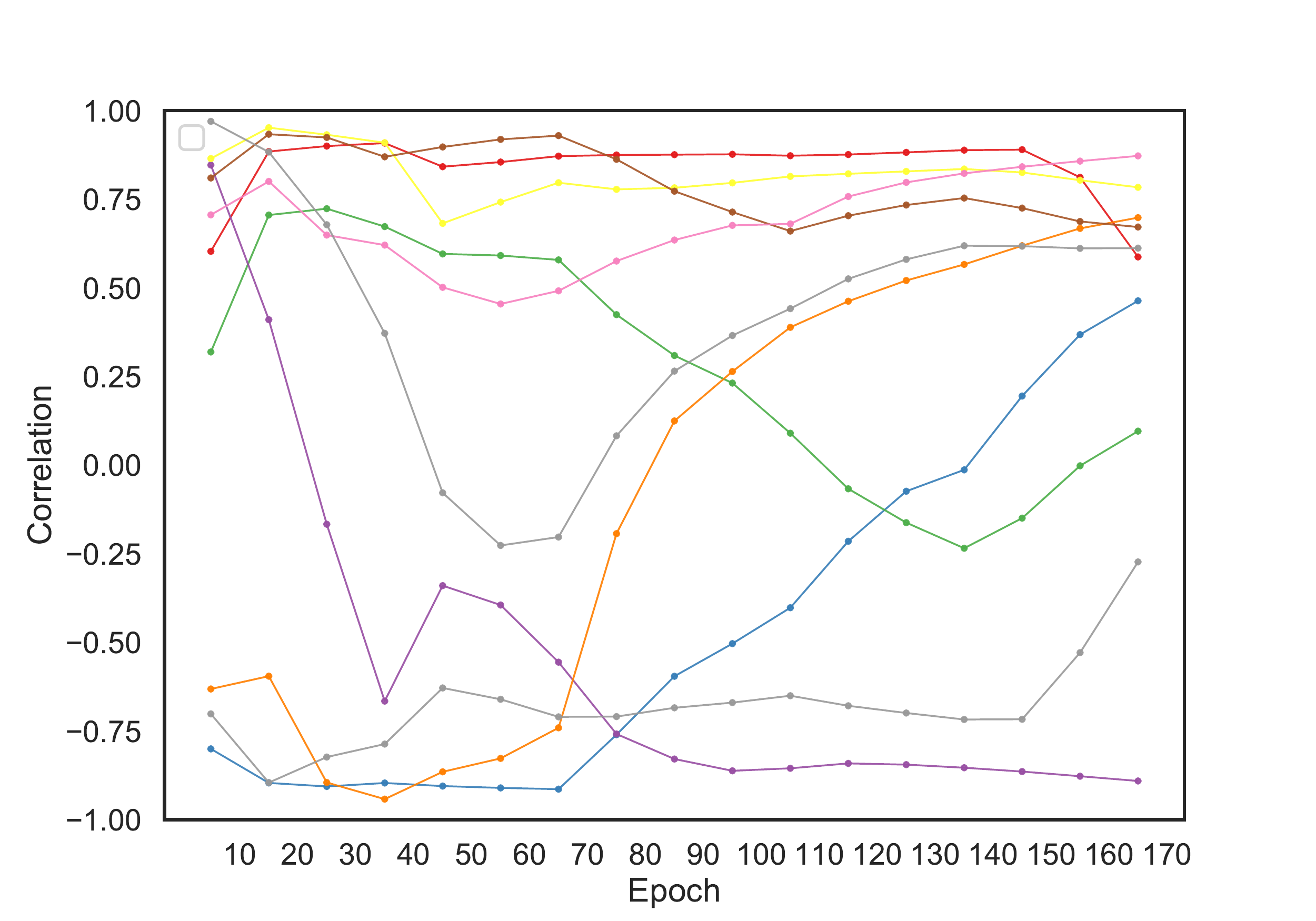}
  \caption{The 2nd FC layer}
  \label{fig:The fourth layer}
\end{subfigure}
\caption{The correlation of random pairs of weights in LeNet trained on the MNIST dataset}
\label{weight correlation LeNet}
\end{figure*}

Training samples contribute differently to loss decreases in a training session. Inspired by the data influence discussion in \cite{koh2017understanding}, we build a reward system based on loss updates from different training samples. As it can be difficult to take each training sample as one arm with a large dataset \cite{cook1980characterizations,koh2017understanding}, we consider each training mini-batch as one arm, and every one-step gradient update on a mini-batch is an arm pull action. 

The setting is that in the current (re)training session we select a subset of samples that are going to be used in subsequent training steps. Formally, while training on $D^m \bigcup D^{m+1}$ the goal is to select a subset of samples $S$ and set $D^{m+1}=S$ for training in the next step based on $D^{m+1} \bigcup D^{m+2}$. 
The key idea is to train for a certain number of epochs based on an optimization technique and then to switch to a strategy of selecting a mini-batch in each training step based on MAB or ``simulating'' such a behavior. In the former case a mini-batch is selected based on MAB while in the latter case standard epoch-based training is performed. In each step we record which mini-batch would have been selected if MAB-based training had been employed. The selection of a mini-batch is based on an MAB algorithm. We select the best MAB algorithm by experimenting with all of the previously introduced MAB algorithms in Section \ref{Multi-armed Bandits}. The arms/mini-batches used most often are part of $S$ (for training session $m+2$).

Each arm pull gives a stochastic reward since the weights are different in each pull, and we propose two reward collection methods: 1) the loss change when making a gradient update based on the mini-batch (denoted by MAB-Loss in experiments) and 2) the L2-norm of gradients of the mini-batch (labeled as MAB-NGrad in experiments). The gradient norm strategy is based on importance sampling proposed in \cite{wang2017accelerating}. Given mini-batch $B$ and parameters $\theta$ that have just been updated based on $B$, the reward is defined as $\sum_{i\in B} \left\lVert  \nabla l^{m+1}_i(\theta) \right\rVert_2^2$
where $l^{m+1}_i$ is the loss component of $l^{m+1}$ pertaining to sample $i$. 

The reward of each arm may change when we pull the same arm at a different training step due to the different underlying parameters. We aim to choose the most influential mini-batches during training and use them for the subsequent retraining session. We list our MAB-based memory replay algorithm with respect to reward corresponding to the decrease of loss and simulated MAB in Algorithm \ref{MAB-based memory replay}, where hyperparameter $q$ controls how many epochs we use for warming up weights (in the experiments we label this version as MAB-Sim). We have attempted a version where the selected mini-batch based on MAB is also processed, Steps 6 and 7 are replaced by ``for each remaining training iteration'' and Step 8 by ``processing the recorded mini-batch in Step 7''. This variant is denoted by MAB-Opt in the experimental section.

\begin{algorithm}
\caption{MAB-based memory replay algorithm}
\label{MAB-based memory replay}
\begin{algorithmic}[1]
\STATE {\textbf{Input}: $D^m \bigcup D^{m+1}$}
\STATE {\textbf{Output}: $S \subset D^m \bigcup D^{m+1}$}
\STATE {Perform $q$ epochs using epoch-based loss optimization}
\STATE {Collect the decreases of loss when training on mini-batches in the $q^{th}$ epoch as the initial rewards of corresponding mini-batches}
\FOR{each remaining epoch}
\FOR{each mini-batch $b$}
\STATE {Record which mini-batch (an arm) would be selected based on an MAB algorithm}
\STATE {Conduct a one-step gradient update based on the mini-batch $b$}
\COMMENT{The recorded MAB mini-batch might be different from the processed mini-batch}
\STATE {The reward received of the mini-batch $b$ is the decrease of loss of this mini-batch}
\ENDFOR
\ENDFOR
\STATE {Order all mini-batches based on the number of times they have been selected in Step 7}
\STATE {Select the top mini-batches as $S$}
\STATE{$D^{m+1} = S$}
\end{algorithmic}
\end{algorithm}

\subsection{MAB-based Weight Optimization}
\label{MAB-based Weight Optimization}

Training in model serving must be performed quickly since inference on a ``stale'' model is dangerous. One solution to expedite optimization is to train only on a subset of weights at a time. We propose a novel way of updating weights during retraining sessions. The weights are first clustered, and then an MAB algorithm selects one cluster at a time. In this context an arm corresponds to a cluster.  

\subsubsection{Weight Clustering}
We have found that a noticeable number of weights in each layer have strong correlations. We have analyzed the weight values and their pairwise correlations in each layer among different epochs. Figure \ref{weight correlation LeNet} shows an example of pairwise correlations of 10 random weight pairs in each layer of the LeNet model that is trained on the MNIST dataset \footnote{http://yann.lecun.com/exdb/mnist/}. The subplots correspond to the first convolutional (CNN) layer, the second CNN layer, the first fully-connected (FC) layer, and the second FC layer. We compute the Pearson correlation of the pairs for every 10 consecutive epochs. The figure shows that many pairs of weights move in tandem, and there are many pairs with a correlation close to 1. Furthermore, the correlation values are fairly stable with only a few abrupt changes during training. Optimizing over a set of weights that converge in sync should be efficient. 

\begin{figure*}[pt]
\begin{subfigure}[b]{0.5\textwidth}
  \includegraphics[width = \textwidth]{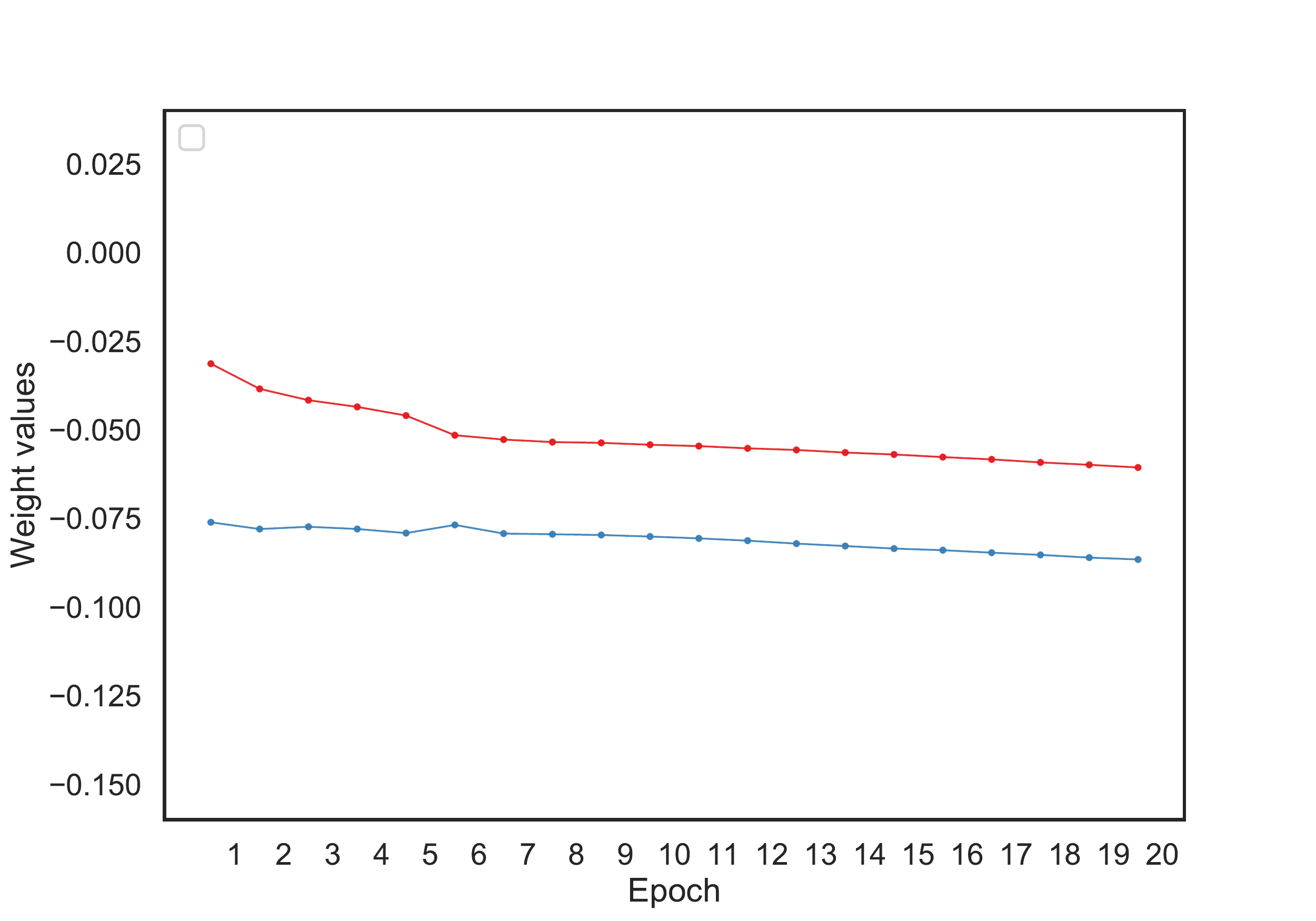}
  \caption{Weight values}
  \label{fig:Weight values a}
\end{subfigure}%
\begin{subfigure}[b]{0.5\textwidth}
  \includegraphics[width =\textwidth]{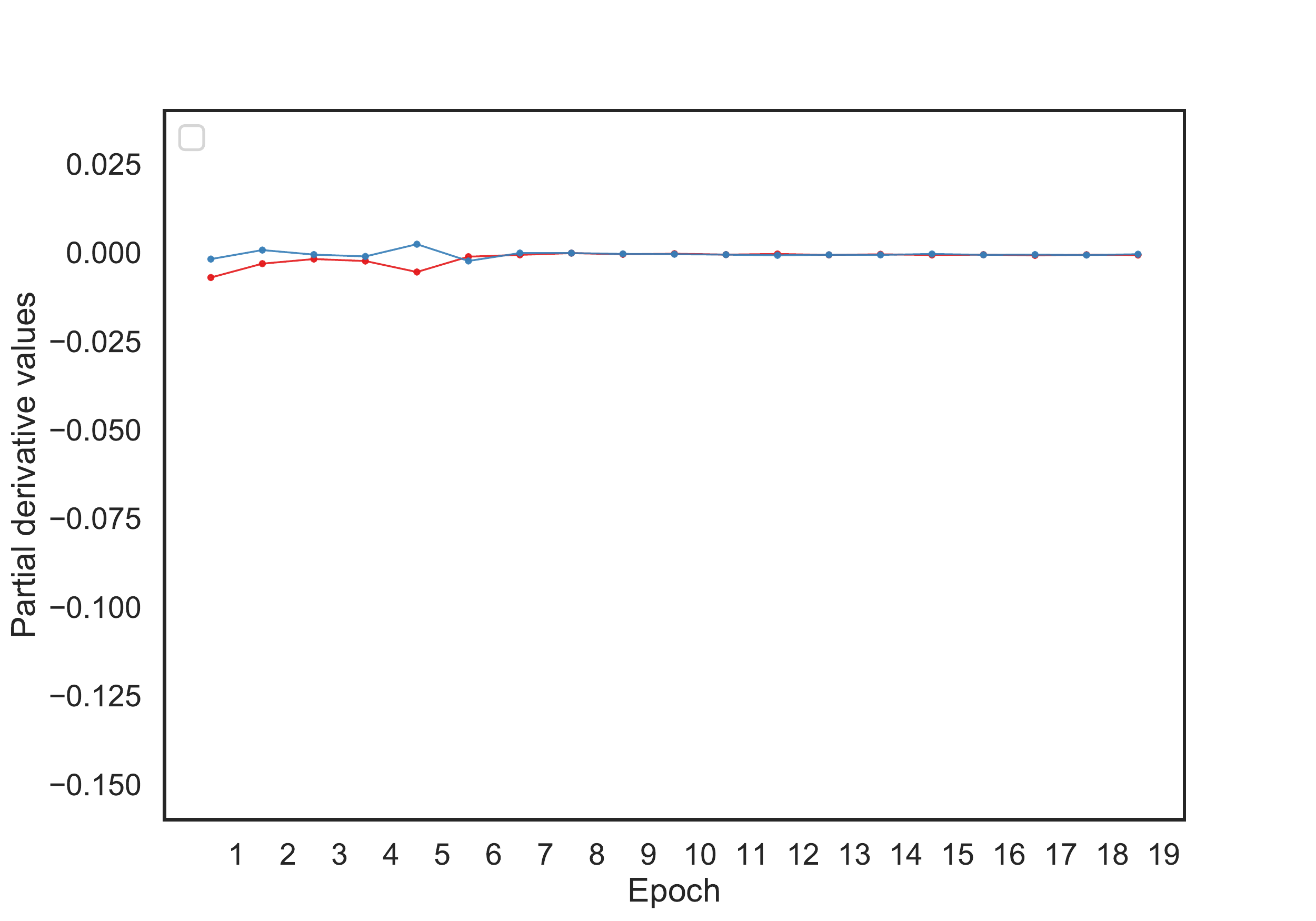}
  \caption{Partial derivatives}
  \label{fig:partial derivatives a}
\end{subfigure}%

\begin{subfigure}[b]{0.5\textwidth}
  \includegraphics[width = \textwidth]{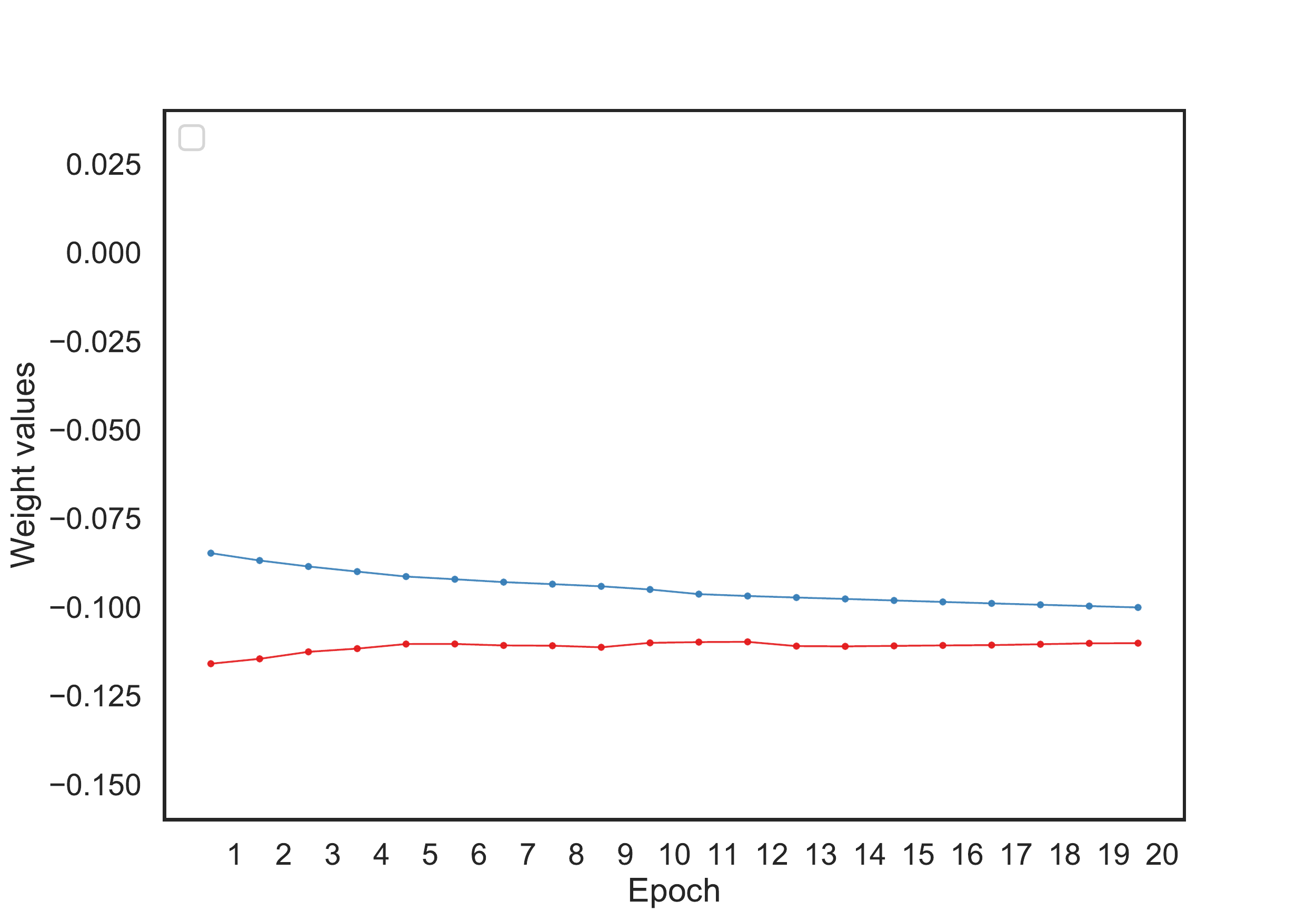}
  \caption{Weight values}
  \label{fig:Weight values b}
\end{subfigure}%
\begin{subfigure}[b]{0.5\textwidth}
  \includegraphics[width =\textwidth]{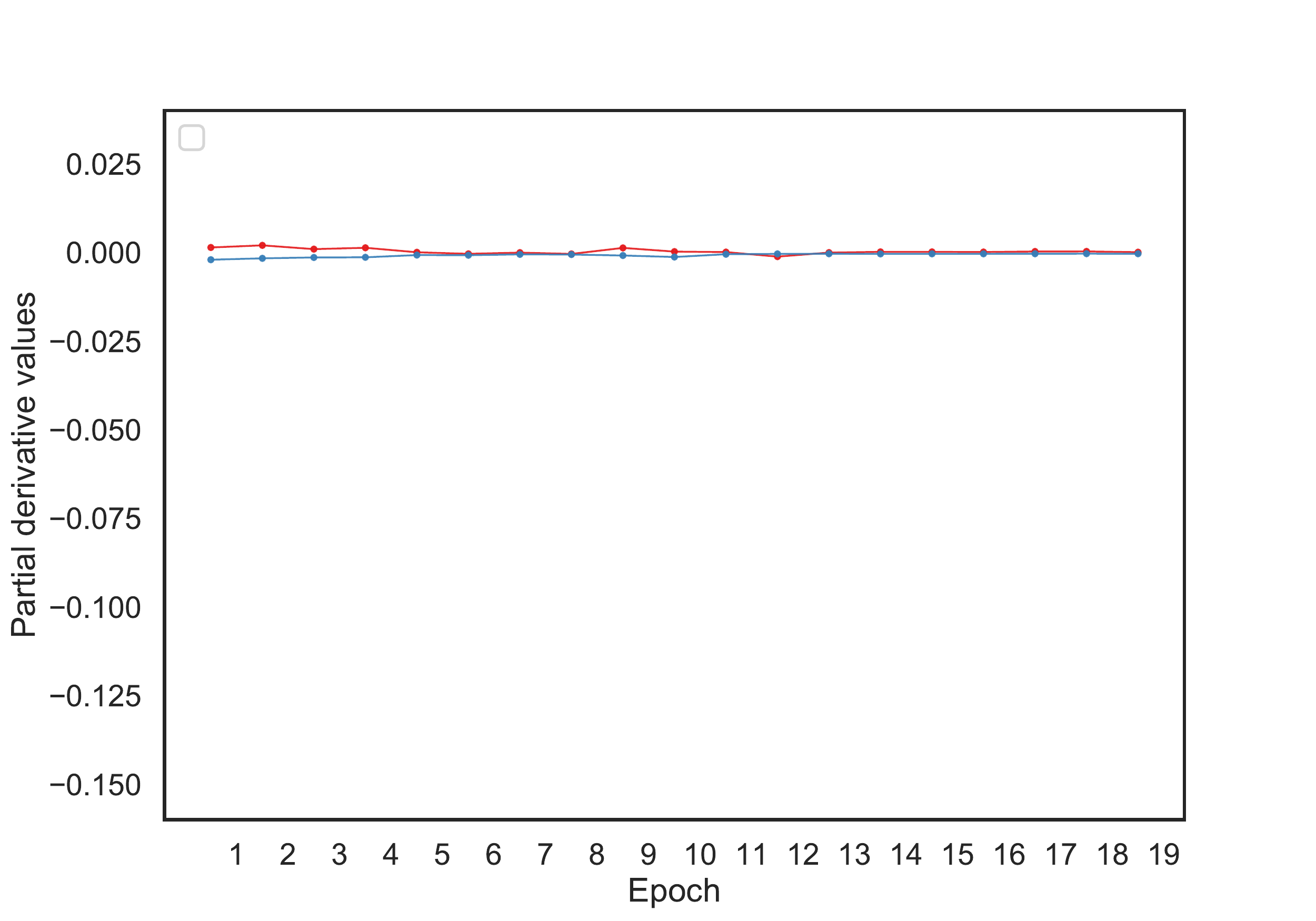}
  \caption{Partial derivatives}
  \label{fig:partial derivatives b}
\end{subfigure}
\caption{The same pair of weights using different weight initialization seeds: Figures \ref{fig:Weight values a} and \ref{fig:partial derivatives a} use one seed, and Figures \ref{fig:Weight values a} and \ref{fig:partial derivatives a} use a different seed}
\label{weight pairs and gradient pairs}
\end{figure*}

We demonstrate the weight and partial derivative relationships of one weight pair in the first layer of the LeNet model in Figure \ref{weight pairs and gradient pairs}. Figures \ref{fig:Weight values a} and \ref{fig:partial derivatives a} illustrate the value series and the partial derivative series of the pair. Figures \ref{fig:Weight values b} and \ref{fig:partial derivatives b} also illustrate the series of the same pair but using a different weight initialization seed. By comparing Figures \ref{fig:Weight values a} and \ref{fig:Weight values b}, the two weights each end up with different values when they are close to convergence. Nevertheless, in later epochs they have a strong correlation. This correlation relationship can be easily verified in Figures \ref{fig:partial derivatives a} and \ref{fig:partial derivatives b} as the partial derivatives of the two weights become close regardless of the initial weights. Ideally, all weights should have gradients close to 0 when a model converges. However, as most deep learning tasks are non-convex problems, not all weights converge at the same rate \cite{ge2015escaping}. 

The novelty of our weight clustering method is that we cluster weights that converge in sync and retrain them together. One option is to cluster the weights based on correlation but in such a case a distance-based algorithm must be used which does not scale. In order to capture trends in weights, we do not use weight values as features but the change in a weight value in two consecutive epochs. We select the values in the last 20\% of the epochs and use standard Euclidean distance as the distance measure in clustering. We have attempted K-Means and DBSCAN clustering algorithms to cluster the weights in each layer with the former performing better. 

We obtain the final clusters of all weights as follows. If the largest number of clusters in different layers is $K$, we create $K$ arms/clusters. For cluster $i,1\le i\le K$, we select a random cluster from each layer and cluster $i$ is the union of all such sets. Those clusters at layers that have already been selected, are not selected for subsequent clusters (it is possible that some layers end up with no clusters to select from in subsequent iterations). Note that, for example, cluster $K$ could consist of only a cluster of a single layer (the layer with the largest number of clusters).

Although in network compression or network pruning studies \cite{saito2007bidirectional,han2015deep_compression,kilinc2017auto,wu2018deep} weights are also clustered for reducing the number of trainable parameters in the retraining phase, the differences are two-fold compared to our weight clustering. In compression and pruning, first, weights are clustered based on their values and not the difference in weight values in two consecutive epochs. Second, in pruning only cluster centroids are trained in the retraining phase, and the rest of the weights are discarded. In our context all of the weights are used in subsequent MAB-based optimization described in the next section. We do not discard any weights but only strategically update a cluster of weights in each mini-batch of our retraining phase. The common number of clusters per layer in our experiments is 3 to 20 obtained by measuring performance on the validation dataset. Next, we explain how we efficiently re-optimize weights using an MAB algorithm during retraining.

\begin{algorithm}
\caption{MAB-based retraining with mini-batch updates}\label{retraining}
\begin{algorithmic}[1]
\STATE {Cluster weights in each layer with respect to $\theta^m$}
\FOR{each cluster $C$}
\STATE{$\theta = \theta^m$}
\STATE {Perform one epoch to optimize only weights in $C$ (freeze other weights)}
\STATE {Collect the decrease of loss of only optimizing this cluster on one epoch as the initial reward of this cluster}
\ENDFOR
\STATE{$\theta = \theta^m$}
\FOR{each epoch training on $D^{m} \bigcup D^{m+1}$}
\FOR{each mini-batch}
\STATE {Pull a cluster of weights (an arm) $C$ using an MAB algorithm}
\STATE {Optimize only weights in $C$ (freeze other weights) and this mini-batch to update $\theta$}
\STATE {Collect the decrease of loss as the reward of this arm}
\STATE {Update the average reward of selecting this arm based on its number of selections and the new reward}
\ENDFOR
\ENDFOR
\end{algorithmic}
\end{algorithm}

\subsubsection{Dynamic Weight Optimization Using Multi-armed Bandits}
In this section, we explain the overall optimization algorithm. In a retraining session, we first iterate over each cluster and only optimize this cluster's set of weights using a single epoch (all data for this retraining session). We collect the loss decrease of pulling each arm/cluster as the initial reward for this arm. For each cluster the initial weights are reset to the initial values. After this initialization step, the mini-batches are processed in the usual epoch-based fashion. 
For each mini-batch, we pull an arm/cluster of weights using an MAB algorithm, collect the loss decrease of optimizing only this arm (the rest of the arms are unchanged), and update the average reward for the selected arm.
The weights are now being updated, but only the weights pertaining to the current mini-batch and arm/cluster are being changed. 
We summarize this retraining strategy MAB-MiniB in Algorithm \ref{retraining}.

An alternative strategy is to optimize over epochs by switching the order of Line 9 and Line 10 of Algorithm \ref{retraining}. This version is called ``Epoch." We pull one arm/cluster of weights at the beginning of every epoch and update only the weights in the selected cluster for all of the mini-batches in that epoch. The reward corresponds to the loss decrease or the L2-norm of gradients of the entire cluster.

We experiment with all aforementioned popular MAB algorithms and choose the best one for each one of the MAB-MiniB and MAB-Epochs algorithms. 
During each mini-batch training, only one cluster/subset of weights receive gradient updates, but all weights are optimized overall in a retraining session. As opposed to dropout--where weights are randomly dropped--our retraining methodology strategically decides which weights receive gradient updates when training a mini-batch.

\section{Experiments and Results}
\label{4-Experiments}
In this section, we introduce the datasets we use, the different types of neural networks, and the experimental setup. In addition to the model retraining experiments, we also demonstrate the generalization effects of combining weight clustering and MAB-based weight optimization.

\subsection{Model Retraining}

\subsubsection{Datasets and Experimental Setting}
\label{4-Datasets}
In order to simulate training a model with a continuous flow of new data, we create the following retraining setting. Given a public dataset, we first randomly partition the data into 6 sets (one set of 50\% and the remaining sets of 10\% each), then we further split each one of the sets into 3 subsets: training (70\%), validation (10\%), and test (20\%). This yields training data $TR$, $R_1$, $R_2$, ..., and $R_5$, validation data $VA$, $A_1$, ..., and $A_5$, and test data $TE$, $E_1$, ..., and $E_5$. We use $TR,VA,TE$ for initial training while each $R_i,A_i,T_i$ represents new data for retraining session $i$. 
When a new batch $i$ of data is received in model serving, we execute retraining of session $i$; the algorithms from Section \ref{MAB-based Memory Replay} are used to select a subset of $TR \cup R_1 \cup \cdots \cup R_i$ to use as training data. In addition, in retraining session $i$, we use $VA \cup A_1 \cup \cdots \cup A_i$ as the validation dataset, and we employ inference on $TE \cup E_1 \cup \cdots \cup E_i$. It is conceivable to potentially also use $A_i$ as validation, and $E_i$ for test. We choose the former strategy since it offers great variability in data, i.e., robustness.  

The weights that lead to the highest accuracy on the validation dataset for each (re)training session are used for inference on test. We showcase our retraining model with six widely used benchmark datasets. We use two datasets for image classifications: MNIST and CIFAR-10 \cite{krizhevsky2009learning}, two datasets that have feature concept drifts: SEA and ELEC\footnote{https://github.com/vlosing/driftDatasets}, and two datasets for text classifications: IMDB\footnote{https://datasets.imdbws.com/} and REUTERS \footnote{https://archive.ics.uci.edu/ml/datasets/reuters-21578+text+categorization+collection}.

Our methods work with any type of a neural network. For simplicity, we use the LeNet framework for the CIFAR-10 and MNIST datasets, a three-layer perceptron (MLP) network for the SEA and ELEC datasets, and an LSTM model followed by a softmax layer for the IMDB and REUTER datasets.

The LeNet and LSTM models are trained using at most 50 epochs with the first 20 epochs for warming up weights in the MAB-based memory replay algorithm; the MLP models are trained using at most 20 epochs with the first 10 epochs for warming up. All models use the Adam optimizer, and we use early stopping of no accuracy increase on validation of up to $10^{-6}$ in 10 consecutive epochs to avoid over-fitting. Using LeNet on the MNIST and CIFAR-10 datasets does not yield state-of-the-art performance numbers but the gap is not too large. Dropout is not applied during MAB-based weight optimization, as dropout also updates the gradients of a subset of weights. Non-MAB methods are tuned with dropout and batch normalization. In MAB-based weight optimization methods we use the scree plot to determine the number of clusters, which is justified later. 

We compare the neuron consolidation method, denoted as NC, to four benchmark regularization methods: fine-tuning using trained weights (Fine-tune) corresponding to not taking any action, EWC, MAS, and Selfless. We also compare the MAB-based memory replay algorithm to reservoir sampling, a popular memory replay algorithm. 
For each model, we use four data settings: the union of all old and new data (Union), random memory replay (Random-replay), new data only (New-data), and our MAB retraining (MAB). We use the same number of mini-batches in reservoir sampling, the random replay, and the MAB retraining algorithms in every retraining session of each dataset. The samples in the mini-batches for any retraining session occupy 10\% of the total training data captured by the Union setting (same as the ratio of $R_i$ over the total training data). We examine in the next section the impact of this choice. 

In the MAB settings we try different configurations for calculating rewards and weight optimizations in our retraining methodology. We select the best MAB algorithm based on the experiments and integrate it with NC (the choices are EI, EI2, EXP3, EXP4, UCB, and TS).

We list all the different options as follows.

\noindent \textit{MAB-based weight optimization}: (\texttt{miniB}) We pull an arm for every mini-batch based on Algorithm \ref{retraining}.
(\texttt{Epochs}) We change Algorithm \ref{retraining} so that pulling an arm corresponds to selecting a cluster and performing several epochs on the selected cluster. (\texttt{FullEpochs}) Standard weight optimization based on epochs. No clustering and MAB is used. 

\noindent \textit{Reward}: The reward setting applies to both memory replay and weight optimization. 
(\texttt{Loss}) The reward is based on the loss change.
(\texttt{NGrad}) The reward is calculated with respect to the sum of the square of the gradient norm in the mini-batch. 

\noindent \textit{Memory replay}:
(\texttt{Sim}) We follow Algorithm \ref{MAB-based memory replay}.
(\texttt{Opt}) We utilize the best MAB algorithm to select mini-batches for the next and the current retraining sessions. Mini-batches are not evenly iterated over in the current (re)training session as in Algorithm \ref{MAB-based memory replay}. 

\noindent \textit{Gradient strategy}:
(\texttt{Grad}) In optimization we use gradient descent. 
(\texttt{KFAC}) We use K-FAC as the training optimizer in (\ref{neuron consolidation}) to calculate natural gradients.

We denote the algorithms by specifying the appropriate configuration for each option. For example, algorithm MAB-MiniB-Loss-Sim-Grad (NC) encodes all of the alternatives of the underlying algorithm. The alternatives pertain only to MAB options and NC is used for regularization since it works best (this is established in the next section). 

We report the accuracy of the test datasets in each retraining session given the aforementioned comparison settings.

\subsubsection{Model Retraining Results}
\label{4-Results}
We first study the impact of the different memory replay strategies. To this end we consider the 5 different strategies and for each one of them we find the best setting with respect to all other algorithmic choices, e.g., regularization and the underlying optimization. In benchmark algorithms we do not consider NC in order to compare only against previously known strategies.
Likewise, for the strategies developed herein we select the best performer and NC is also an option.
This also implies that we compare our best algorithm with respect to the previously best known algorithm under the different memory replay strategies. 
Since models use most of the data in the union setting, we expect this setting to be an upper bound with respect to the accuracy performance. 

\begin{figure*}[pt]
\begin{subfigure}[b]{0.48\textwidth}
  \includegraphics[width = \textwidth]{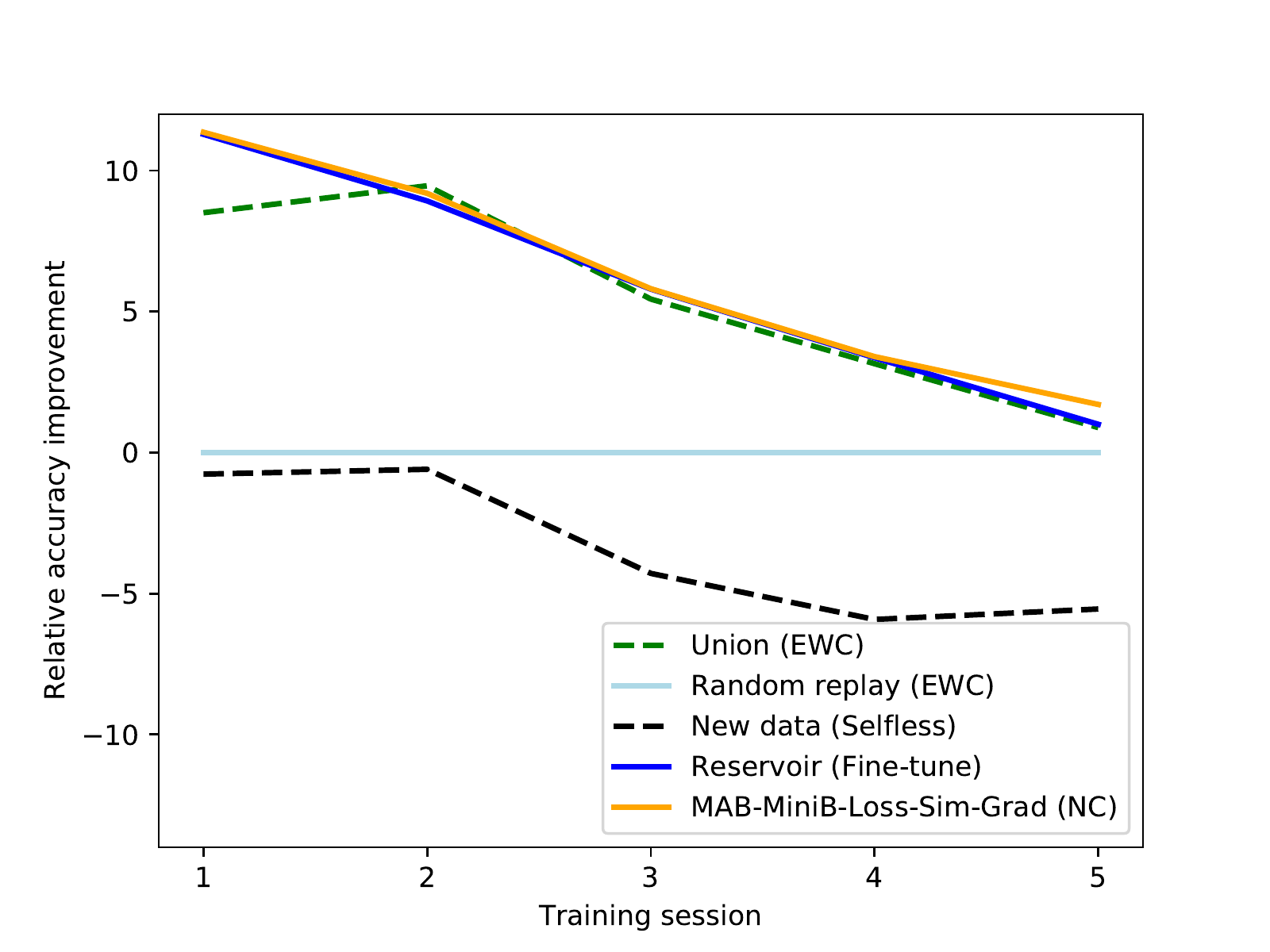}
  \caption{ELEC}
  \label{fig:Best ELEC}
\end{subfigure}%
\begin{subfigure}[b]{0.48\textwidth}
  \includegraphics[width = \textwidth]{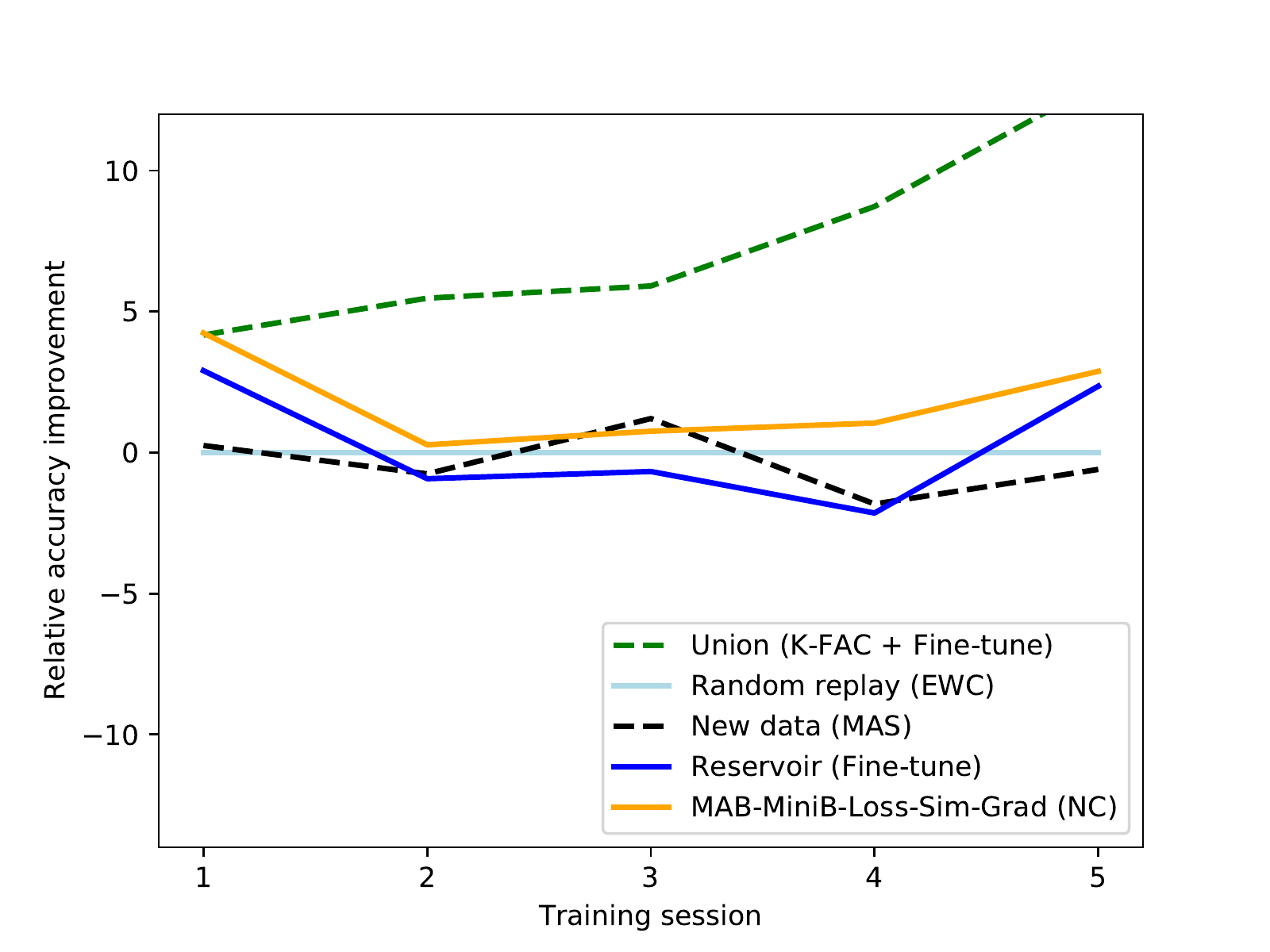}
  \caption{REUTERS}
  \label{fig:Best REUTERS}
\end{subfigure}%

\begin{subfigure}[b]{0.48\textwidth}
  \includegraphics[width = \textwidth]{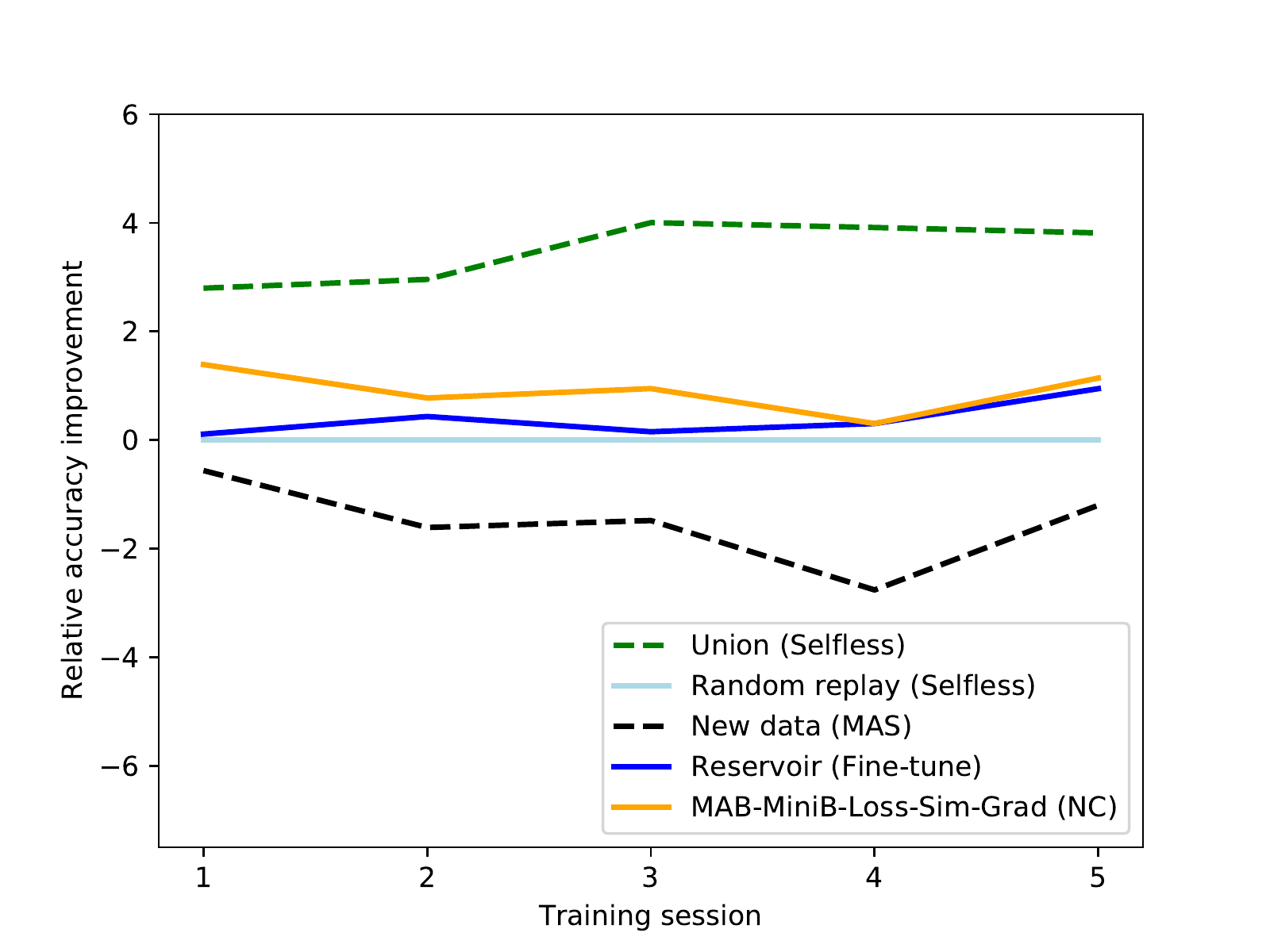}
  \caption{CIFAR-10}
  \label{fig:Best CIFAR-10}
\end{subfigure}%
\begin{subfigure}[b]{0.48\textwidth}
  \includegraphics[width =\textwidth]{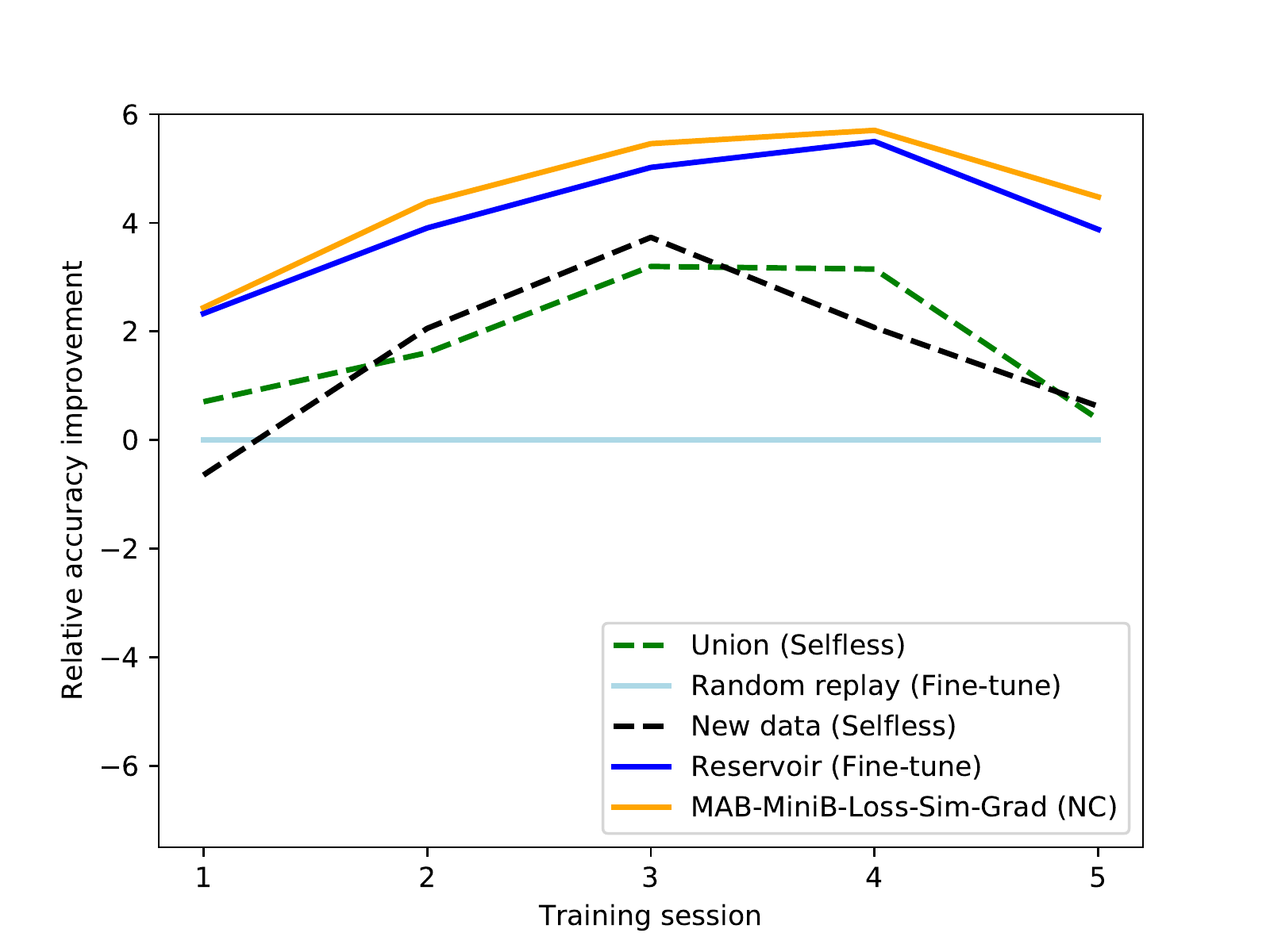}
  \caption{IMDB}
  \label{fig:Best IMDB}
\end{subfigure}%

\begin{subfigure}[b]{0.5\textwidth}
  \includegraphics[width =\textwidth]{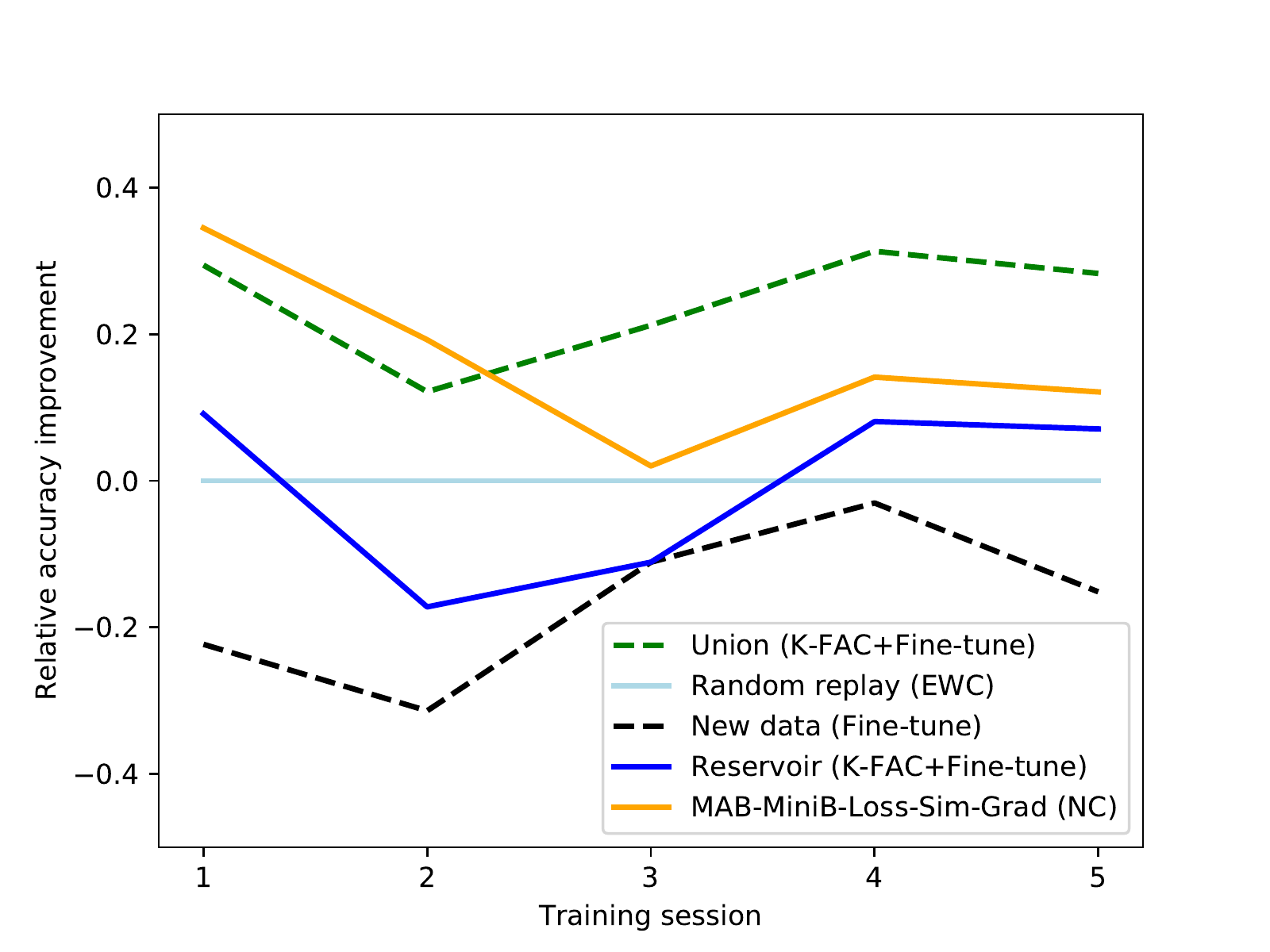}
  \caption{MNIST}
  \label{fig:Best MNIST}
\end{subfigure}%
\begin{subfigure}[b]{0.5\textwidth}
  \includegraphics[width = \textwidth]{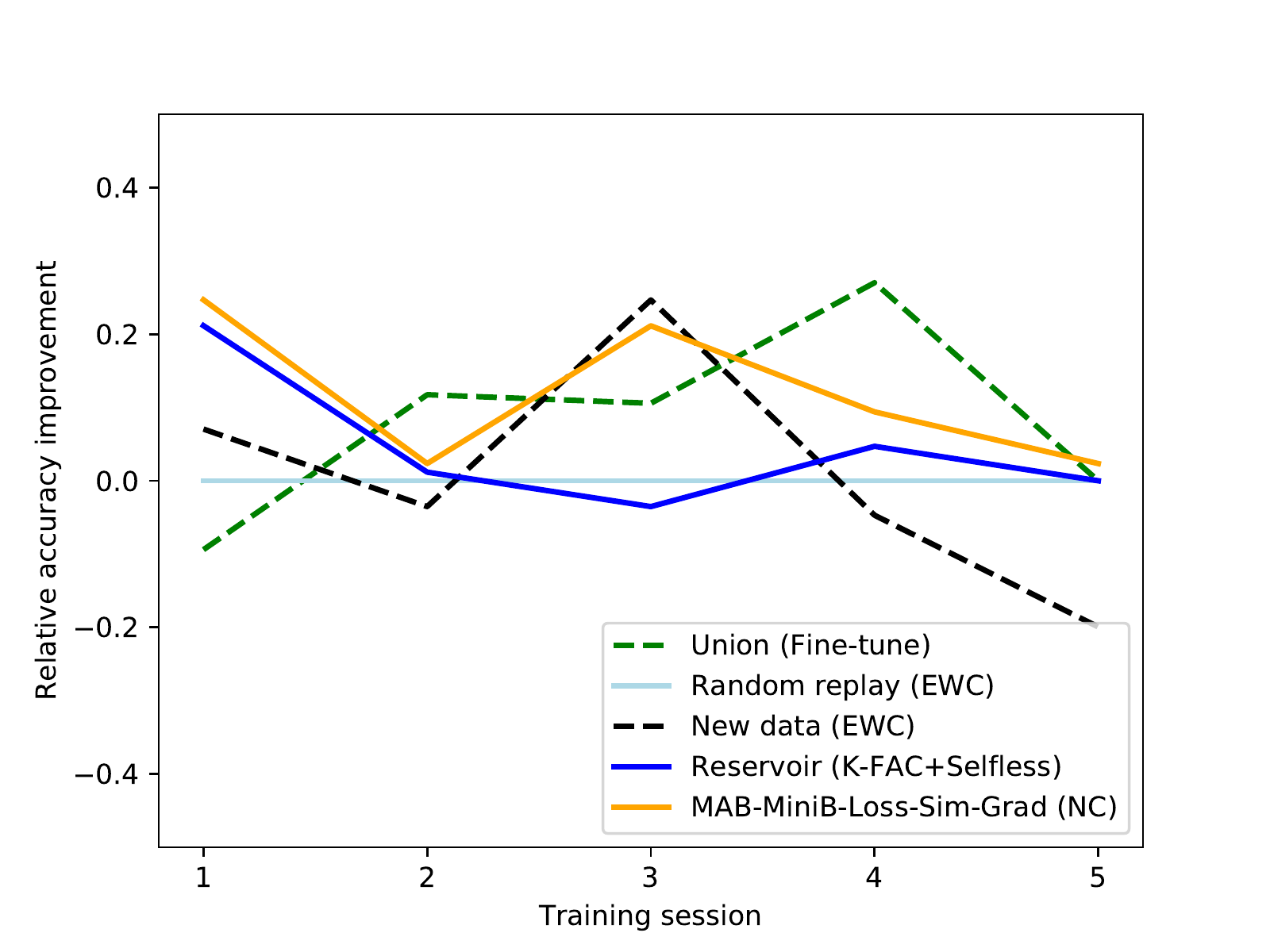}
  \caption{SEA}
  \label{fig:Best SEA}
\end{subfigure}%

\caption{The relative accuracy improvements (\%) of the best MAB retraining model MAB-MiniB-Loss-Sim-Grad (NC), the best benchmark models under the union setting, the random replay setting, the new data setting, and reservoir sampling over the best model under the random replay setting}
\label{Best-benchmark}
\end{figure*}

Table \ref{Best average accuracy}  and Figure \ref{Best-benchmark} compare accuracy under the best MAB setting to the best benchmark model results under the union setting, the random replay setting, the new data setting, and reservoir sampling. 
In the table the numbers in bold present the best performer while the underlined numbers correspond to the second best algorithm; they are the averages across all 5 retraining sessions. Figure \ref{Best-benchmark} breaks down the numbers by session and it also specifies the underlying algorithmic strategy. The table reveals that in 3 datasets MAB outperforms all other models, including the best Union setting. For CIFAR-10 and REUTERS the latter is best, however MAB outperforms all of the remaining models. Union is much more computationally demanding, which is going to be established later; thus we claim that MAB is very robust and it is the algorithm of choice. 

In Figure \ref{Best-benchmark}, we illustrate the trends of the relative improvements of the best union results, the best new data results, the best reservoir sampling results, and the best MAB results over the best random replay results in the five retraining sessions of the six datasets. The improvements achieved by  MAB-MiniB-Loss-Sim-Grad (NC) indicate the performance boost of Algorithms \ref{MAB-based memory replay} and \ref{retraining}. 
The best MAB sampling algorithms corresponding to the six datasets are EI2, EI2, EXP3, TS, EXP3, and EI respectively.
We observe that only New-data and Union sometimes outperform the MAB strategy. 
The numbers in Table \ref{Best average accuracy} are average accuracies over the 5 sessions shown in Figure \ref{Best-benchmark}. 
The integrated MAB retraining model has better performances than the best random replay and reservoir sampling models in most sessions and datasets. In particular, the MAB retraining model sometimes performs better than the best models under the union setting. Because the SEA and the ELEC datasets have concept drifts, the union data setting does not always outperform the memory replay setting or even the new data setting (the drift likely lingers in the union setting even after a random creation of sessions). The difference in performance between the MAB retraining setting and the union setting for the REUTERS dataset is larger than the rest because REUTERS has 46 classes which essentially require a large amount of old data for retraining. 

In Table \ref{Best relative accuracy}, we show the average relative accuracy improvements of the best MAB model MAB-MiniB-Loss-Sim-Grad (NC) over the best benchmark models for the six datasets (we divide by MAB-MiniB-Loss-Sim-Grad (NC)). The models correspond to the models in Figure \ref{Best-benchmark} and Table \ref{Best average accuracy}. Positive values reveal that MAB-MiniB-Loss-Sim-Grad (NC) outperforms. Union is the best performer with a much higher computational time, however MAB outperforms all other choices including reservoir, which is deemed state-of-the-art. The overall improvement with respect to reservoir is 0.48\%. 

\begin{table}[h!]
\centering
\scalebox{0.7}{
\begin{tabular}{|l|r|r|r|r|r|r|}\hline
& CIFAR-10 & MNIST  & SEA  & ELEC  & IMDB & REUTERS \\ \hline
Union                   &\textbf{67.59}     &\textbf{99.13}           &85.23           &77.10           &84.45  &\textbf{62.42}\\\hline
Random-replay           &65.30           &98.89           &85.17           &73.17           &82.96           &58.06\\\hline
New-data                &64.30           &98.72           &\underline{85.25}           &70.60           &84.05           &57.85\\\hline
Reservoir               &65.56           &98.88           &85.21           &\underline{77.51}           &\underline{86.37}           &58.22\\\hline
MAB                     &\underline{65.79}           &\underline{99.05}  &\textbf{85.27}  &\textbf{77.67}  &\textbf{86.68}           &\underline{59.11}\\\hline
\end{tabular}}
\caption{Best average accuracy (\%) for the best MAB retraining model (MAB-MiniB-Loss-Sim-Grad (NC)) denoted as MAB, and the best benchmark models under different training settings for the six datasets}
\label{Best average accuracy}
\end{table}

\begin{table}[h!]
\centering
\scalebox{0.7}{
\begin{tabular}{|l|r|r|r|r|r|r|}\hline
& CIFAR-10 & MNIST  & SEA  & ELEC  & IMDB & REUTERS \\ \hline
Union                   &-2.50          &-0.08           &0.05            &0.74            &2.64             &-5.30\\\hline
Random-replay           &0.92           &0.16            &0.12            &6.15            &4.48             &1.81\\\hline
New-data                &2.49           &0.33            &0.12            &10.01           &2.88             &2.18\\\hline
Reservoir               &0.52           &0.17            &0.07             &0.21           &0.36             &1.53\\\hline
\end{tabular}}
\caption{Average accuracy (\%) improvements of the best MAB retraining model over the best benchmark models under different training settings for the six datasets}
\label{Best relative accuracy}
\end{table}

\begin{figure*}[pt]
\centering
\begin{subfigure}[b]{0.48\textwidth}
  \includegraphics[width =\textwidth]{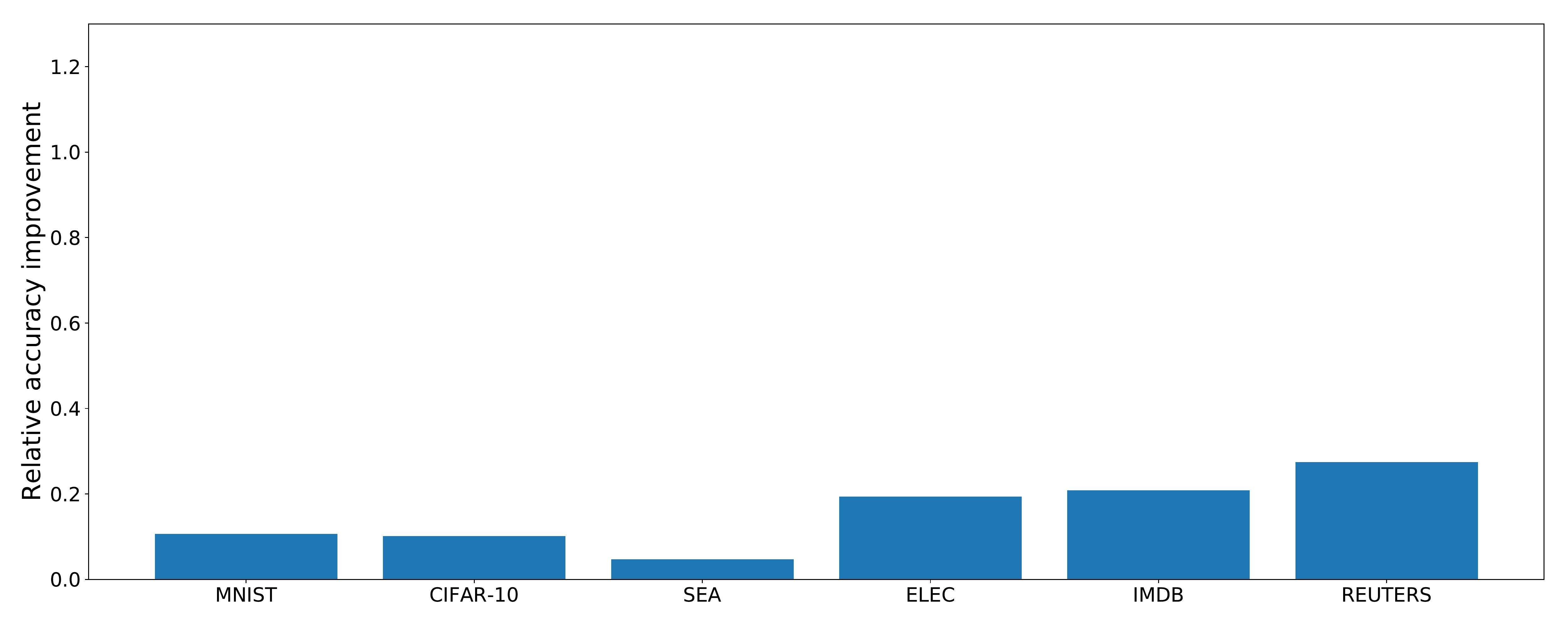}
  \caption{Gains from MAB-based memory reply exclusively}
  \label{MAB-FullEpochs-Loss-Sim-Grad_Reservoir}
\end{subfigure}%
\begin{subfigure}[b]{0.48\textwidth}
  \includegraphics[width = \textwidth]{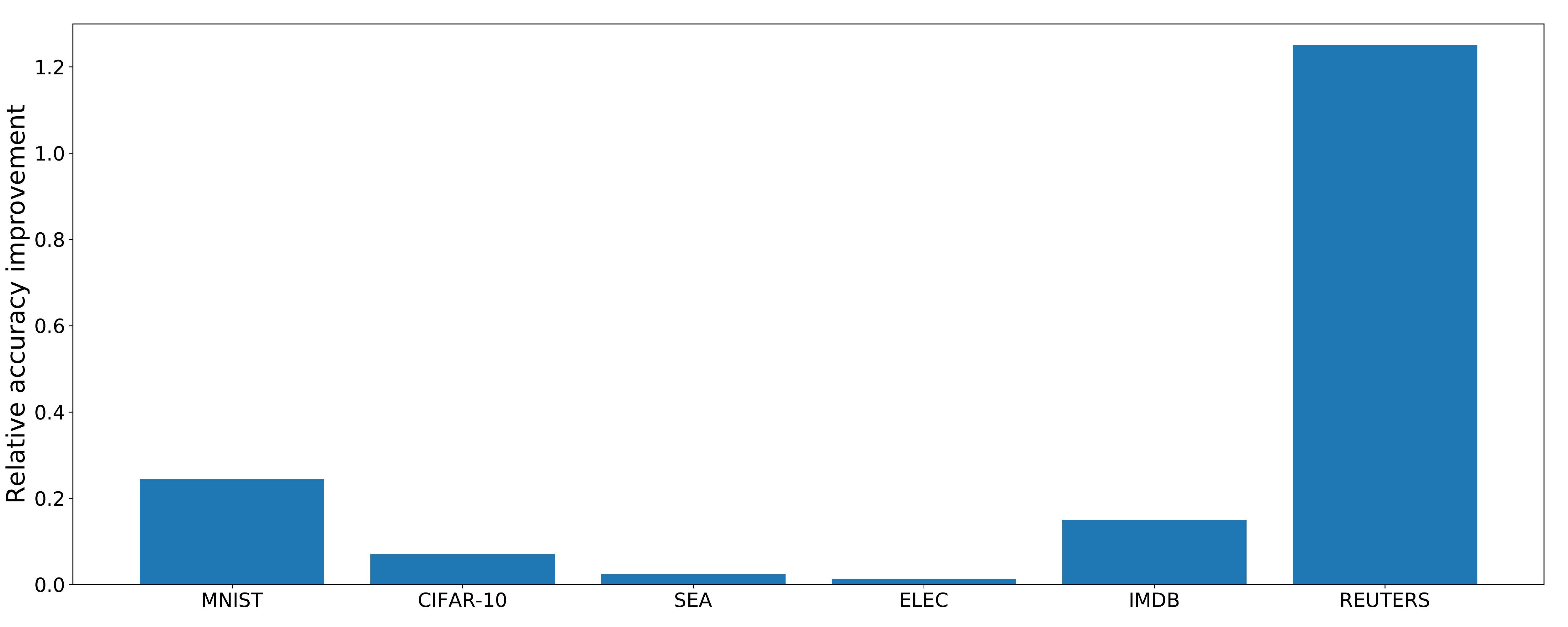}
  \caption{Gains for MAB-based weight optimization}
  \label{MAB-MiniB-Loss-Sim-Grad_MAB-FullEpochs-Loss-Sim-Grad}
\end{subfigure}

\caption{Impact in isolation of MAB-based memory replay and MAB-based weight optimization}
\label{ablation-chart}
\end{figure*}

In order to isolate the impact of the MAB algorithm for weight optimization, we consider MAB-FullEpochs-Loss-Sim-Grad (NC) versus reservoir, which uses the same full epochs approach (we divide by the latter). Note that in this setting the only difference is MAB-based memory replay exhibited in Algorithm \ref{MAB-based memory replay}. The gaps are shown in the top bar chart in Figure \ref{ablation-chart}. The overall average across all numbers is 0.13\%. In order to assess only the impact of MAB-based weight optimization, we examine the gap between MAB-MiniB-Loss-Sim-Grad (NC) and 
MAB-FullEpochs-Loss-Sim-Grad (NC) (we divide by the latter). These algorithms use the same memory replay algorithm, and they only differ in weight optimization. The results are shown in the bottom bar chart in Figure \ref{ablation-chart}. The overall average gap is 0.29\% which demonstrates the efficacy of Algorithm \ref{retraining}. 

\begin{figure*}[pt]
\includegraphics[width =\textwidth]{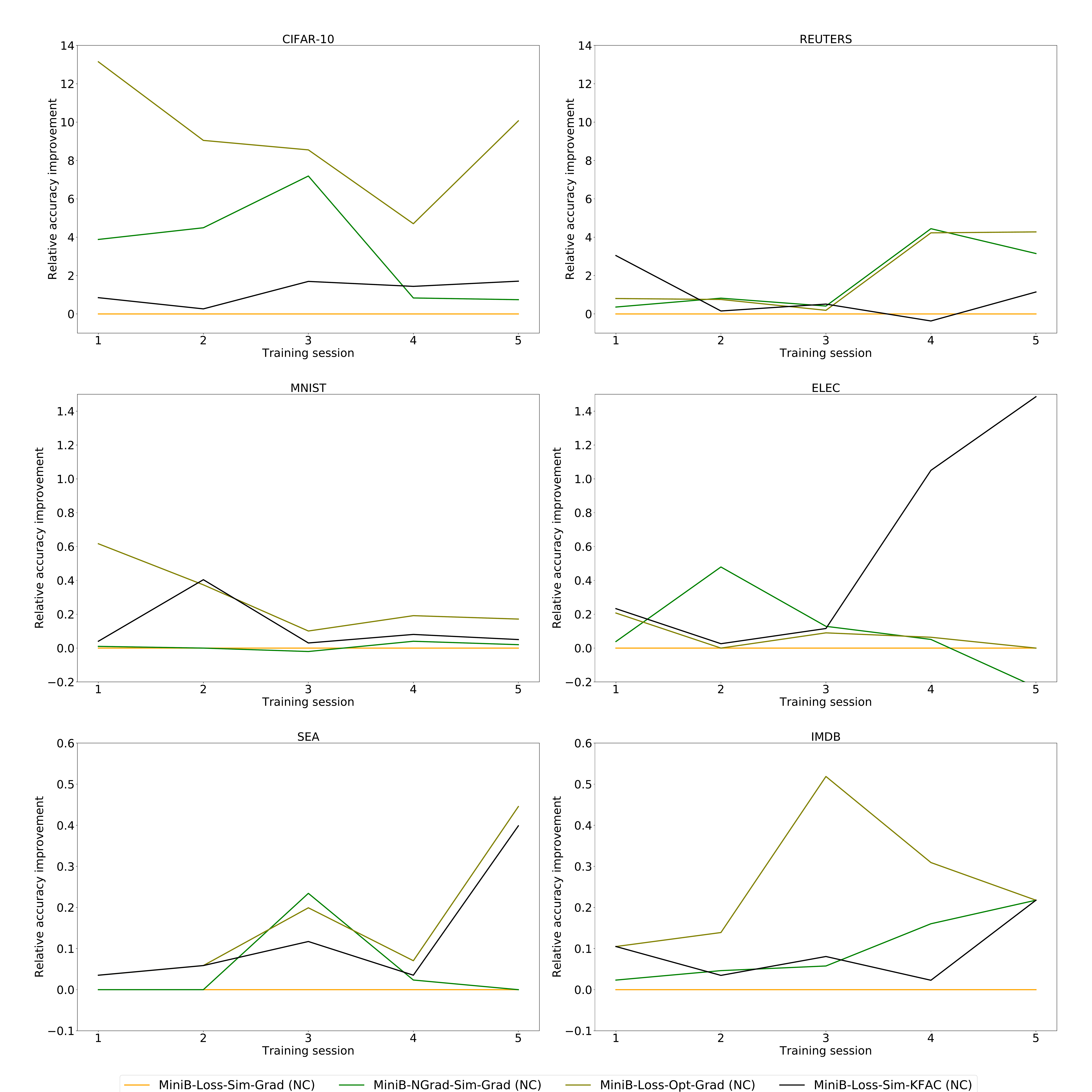}
\caption{The relative accuracy improvements (\%) of MAB-MiniB-Loss-Sim-Grad (NC) over the NC regularization term with other competitive MAB configurations (MAB- is omitted in the legend)}
\label{Best-MAB}
\end{figure*}


Figure \ref{Best-MAB} illustrates the relative improvements of MAB-MiniB-Loss-Sim-Grad over the other MAB configurations (addressed in Section \ref{4-Datasets}) for the six datasets. 
Starting with MAB-MiniB-Loss-Sim-Grad we vary other options one by one. 
The MAB-Epochs algorithms take the full dataset for optimizing each cluster of weights. However, as we do not (re)train a model using many epochs, which leads to a small number of arm pulls, we do not observe superior results compared to MAB-MiniB-Loss-Sim-Grad. MAB-MiniB-Loss-Sim-Grad and MAB-MiniB-NGrad-Sim-Grad have a similar performance indicating that the reward function setting does not have a huge impact. MAB-MiniB-Loss-Sim-KFAC is also competitive in many datasets and sessions, however it also performs very poorly in some situations (ELEC). In general, we observe that the performance of KFAC is very unstable. The performance of MAB-MiniB-Loss-Opt-Grad is the worst among the 5 considered, which leads to the conclusion that performing MAB optimization for selection of mini-batch is not a good strategy. 
In the rest of the paper, we abbreviate MAB-MiniB-Loss-Sim-Grad simply as MAB.

Because neuron regularization has terms for both neurons and weights, it is expected to be more computationally demanding than EWC, MAS, and Fine-tune. Compared to Selfless in the union setting, which also considers both neuron and weight importance, NC regularization in the union setting is 18-22 times faster than Selfless (measured on a 2080 Ti GPU) across the six datasets. The reduction in the time comes from the fact that NC has only individual neuron level terms while Selfless considers pairs of neurons. We find this conclusion universal for all comparison settings and different datasets. 

\begin{figure*}[pt]
\centering
\begin{subfigure}[b]{0.32\textwidth}
  \includegraphics[width =\textwidth]{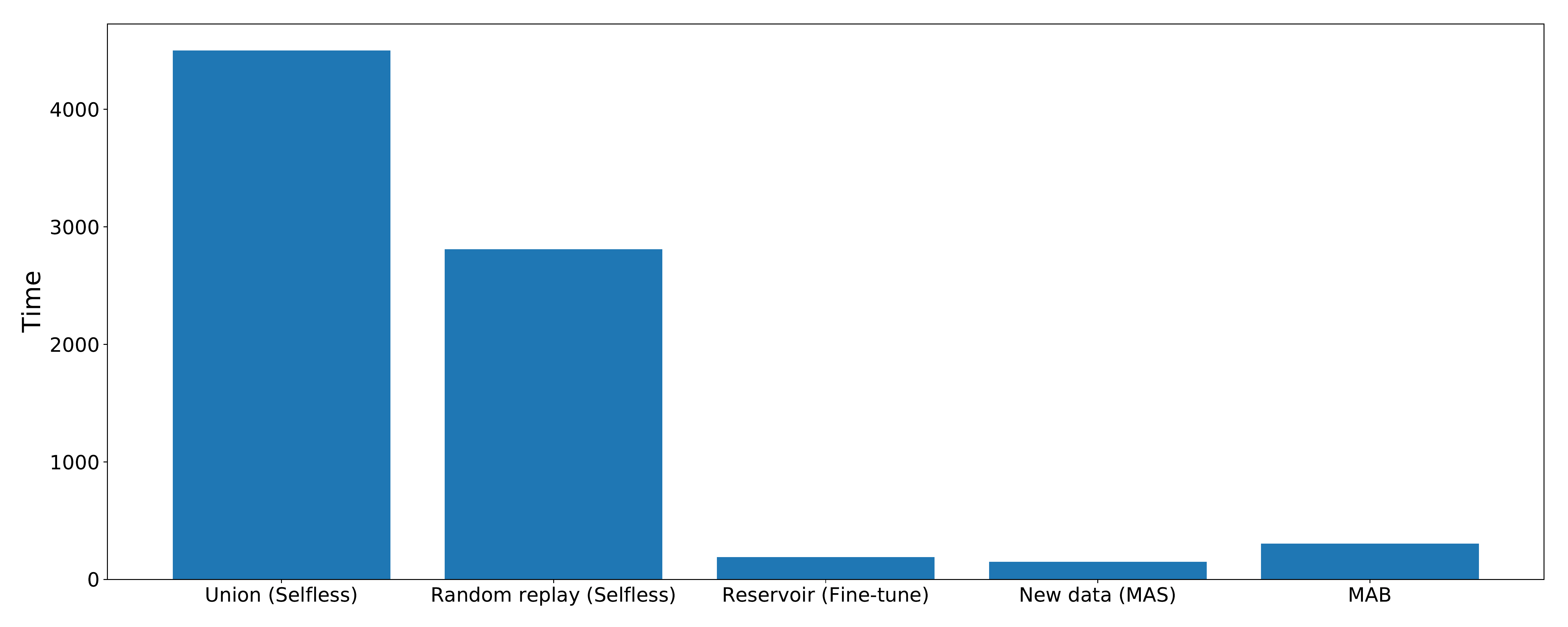}
  \caption{Benchmark models and MAB}
  \label{CIFAR-10 dataset time}
\end{subfigure}%
\begin{subfigure}[b]{0.32\textwidth}
  \includegraphics[width =\textwidth]{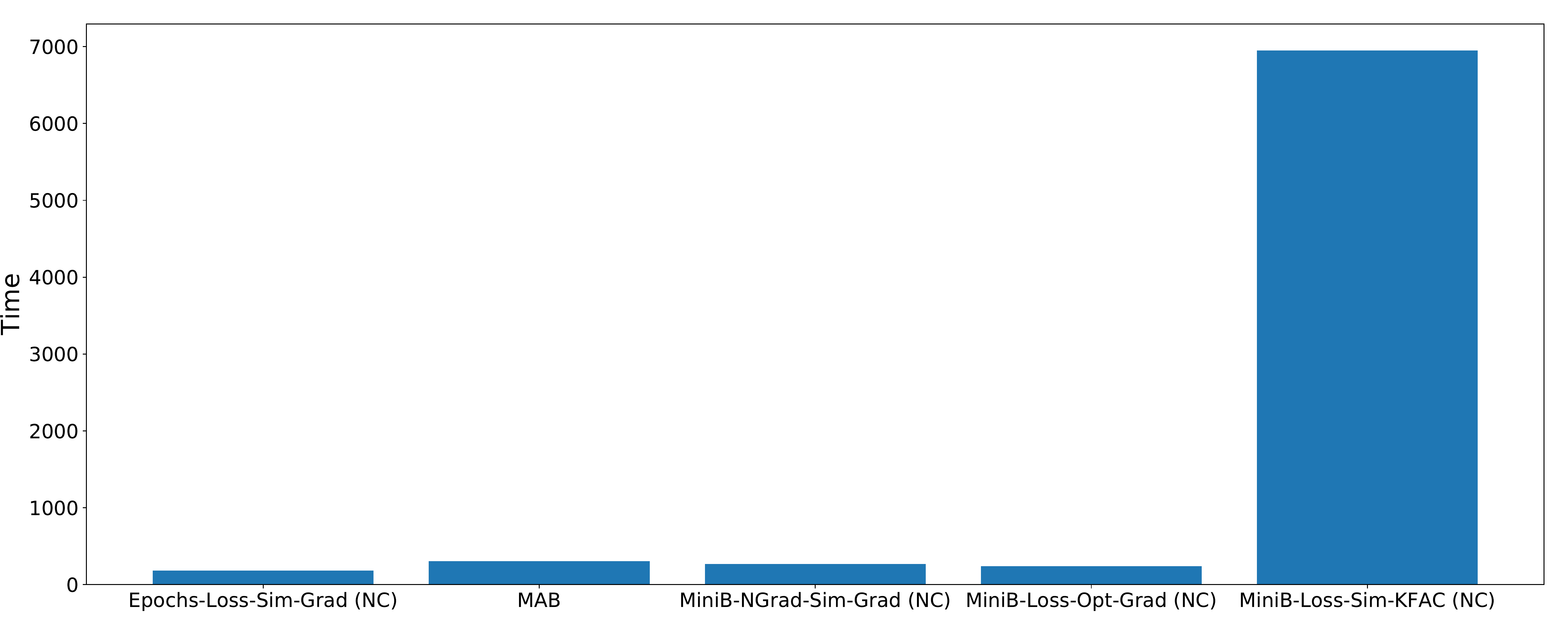}
  \caption{Different MAB configurations}
  \label{CIFAR-10 dataset MAB time}
\end{subfigure}%
\begin{subfigure}[b]{0.32\textwidth}
  \includegraphics[width =\textwidth]{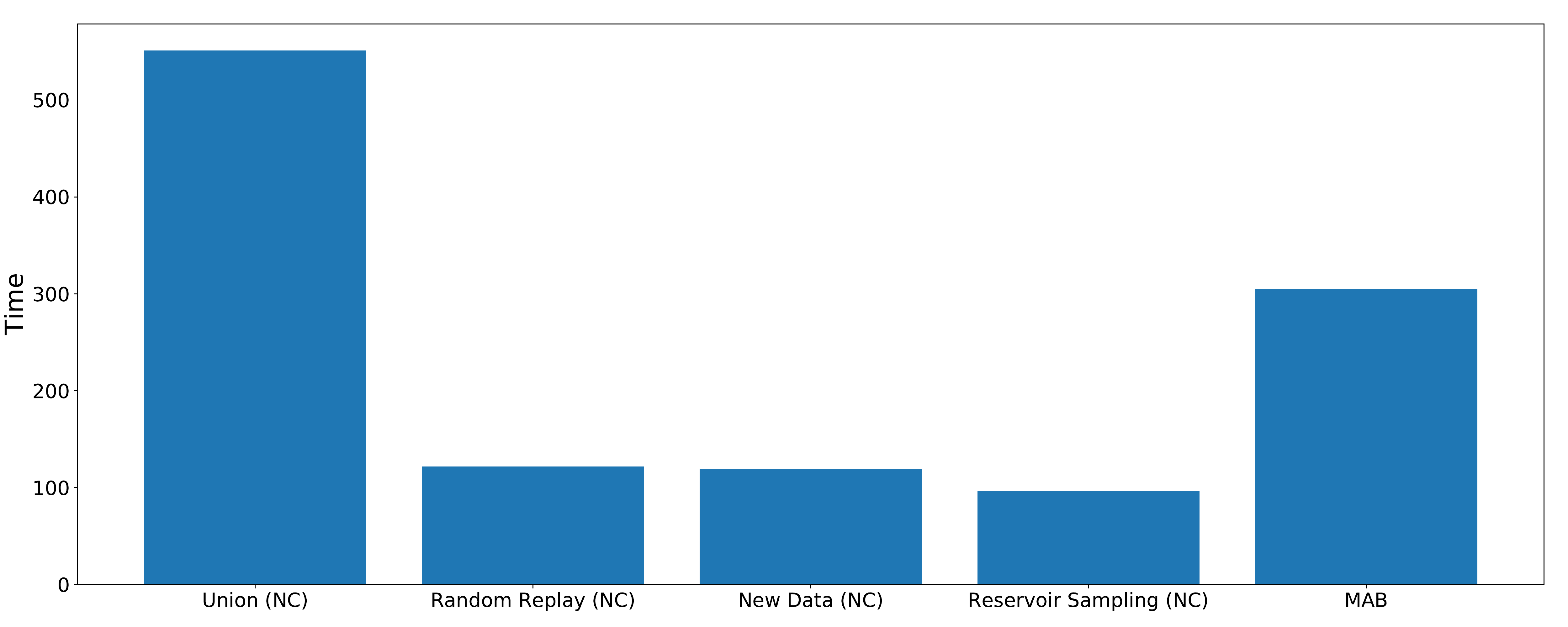}
  \caption{NC under different data settings}
  \label{CIFAR-10 NC time}
\end{subfigure}%

 
 
 
\caption{Average training time (s) on the CIFAR-10 dataset under different settings; In (b) MAB is omitted in labels}
\label{CIFAR-10 time}
\end{figure*}

We compare the average training time of MAB to the best benchmark regularization methods under the union setting, the random replay setting, the new data setting, and reservoir sampling in Figure \ref{CIFAR-10 dataset time}. When training on the same amount of data, random replay with Selfless is 3 times slower than MAB as demonstrated in the top figure. The figure clearly indicates that Selfless is very slow and the remaining three strategies have computational requirements in the same range with MAB being the slowest one.  
We also compare the average training time for different MAB configurations shown in Figure \ref{CIFAR-10 dataset MAB time}. MAB-Epochs-Loss-Sim-Grad has the shortest training time, while MAB-MiniB-Loss-Sim-KFAC is the slowest, which is expected. 
Except for KFAC, the remaining versions exhibit similar model training time. 
Although the union data setting usually yields a better performance compared to MAB-based retraining, the union data setting requires excessive training time and computation resources. Figure \ref{CIFAR-10 NC time} shows the average training time comparison of the union setting, the random replay setting, reservoir sampling, the new data setting, and the best MAB setting using the NC regularization term, which is a superior regularization. Union is clearly the slowest one, as expected, followed by MAB and then the remaining three algorithms. 
MAB is slower than these algorithms despite all of them using the same number of samples due to the extra time to run the actual multi-arm bandit strategy. 

To showcase the robustness of the MAB retraining model against reservoir sampling given different ratios of selected data samples, we detail the average accuracy comparison in Table \ref{average accuracy ratio} and average running time in Table \ref{average time ratio}. In Table \ref{average accuracy ratio} we point out that in every single case MAB outperforms reservoir. The relative accuracy improvements on average for ratios 30\%, 20\%, and 5\% are 0.50\%, 0.47\%, 0.42\%, respectively, while on average the training time of MAB increases by 11.71\%, 17.55\%, and 19.33\%, respectively. By using sparse tensor operations the computational times of MAB can be further improved since MAB is using only on average 25\% of the weights as discussed later. 

\begin{table}[]
\scalebox{0.75}{
\begin{tabular}{l|r|r|r|r|r|r|}
\cline{2-7}
                                                      & \multicolumn{2}{l|}{30\%} & \multicolumn{2}{l|}{20\%} & \multicolumn{2}{l|}{5\%} \\ \cline{2-7} 
                                                      & MAB       & Reservoir     & MAB       & Reservoir     & MAB      & Reservoir     \\ \hline
\multicolumn{1}{|l|}{CIFAR-10}      &66.25	&66.11	&66.16	&65.69	&64.20	&63.77 \\ \hline
\multicolumn{1}{|l|}{MNIST} 	    &99.08	&98.91	&99.08	&98.93	&98.75	&98.68 \\ \hline
\multicolumn{1}{|l|}{SEA} 	        &85.27	&85.21	&85.26	&85.21	&85.26	&85.25 \\ \hline
\multicolumn{1}{|l|}{ELEC} 	        &77.77	&77.30	&77.16	&76.72	&71.44	&71.25 \\ \hline
\multicolumn{1}{|l|}{IMDB} 	        &86.72	&86.45	&86.59	&86.35	&83.14	&82.31 \\ \hline
\multicolumn{1}{|l|}{REUTERS} 	    &60.05	&59.08	&59.89	&59.26	&56.91	&56.66 \\ \hline
\end{tabular}
}
\caption{Average accuracy under different sample ratios}
\label{average accuracy ratio}
\end{table}

\begin{table}[]
\scalebox{0.75}{
\begin{tabular}{l|r|r|r|r|r|r|}
\cline{2-7}
                                                      & \multicolumn{2}{l|}{30\%} & \multicolumn{2}{l|}{20\%} & \multicolumn{2}{l|}{5\%} \\ \cline{2-7} 
                                                      & MAB       & Reservoir     & MAB       & Reservoir     & MAB      & Reservoir     \\ \hline
\multicolumn{1}{|l|}{CIFAR-10} 	&461	&400	&377	&208	&218	&162 \\ \hline
\multicolumn{1}{|l|}{MNIST} 	&272	&416	&154	&311	&99	    &124 \\ \hline
\multicolumn{1}{|l|}{SEA} 	    &85	    &90	    &61	    &82	    &28	    &49 \\ \hline
\multicolumn{1}{|l|}{ELEC} 	    &84	    &81	    &48	    &46	    &18	    &18 \\ \hline
\multicolumn{1}{|l|}{IMDB} 	    &2010	&1972	&1089	&1075	&617	&476 \\ \hline
\multicolumn{1}{|l|}{REUTERS} 	&3669	&3613	&1445	&1406	&1140	&1133 \\ \hline
\end{tabular}
}
\caption{Average training time (s) under different sample ratios}
\label{average time ratio}
\end{table}

\begin{table}[h!]
\centering
\scalebox{0.75}{
\begin{tabular}{|l|r|r|r|r|r|r|}\hline
& CIFAR-10 & MNIST  & SEA  & ELEC  & IMDB & REUTERS \\ \hline
Union                   &4,502               &2,389                &73                &55               &2,354              &13,388\\\hline
Reservoir               &189           		 &705              	   &47                &32          		&512          		&603\\\hline
MAB           			&305           		 &126  			 	   &46                &34  		  		&707  		 		&690\\\hline
\end{tabular}
}
\caption{The average training time of the six datasets}
\label{Best time}
\end{table}


Table \ref{Best time} shows the average training time of the most competitive models used in Figure \ref{Best-benchmark} for the six datasets. Union clearly has by far the worst computational time, which in our opinion does not justify the improvement in accuracy. MAB is slightly slower than reservoir,  but the difference is not large, i.e., they are on the same scale. More importantly, MAB has better accuracy. 

As shown in later figures, MAB during retraining is using on average only 25\% of the weights however since sparse tensors are not handled by our implementation, this potential benefit is not captured in the computational times. We posit that a sparse tensor implementation would bring the computational time of MAB below the time of reservoir. 

\begin{figure*}[pt]
  \includegraphics[width = \textwidth]{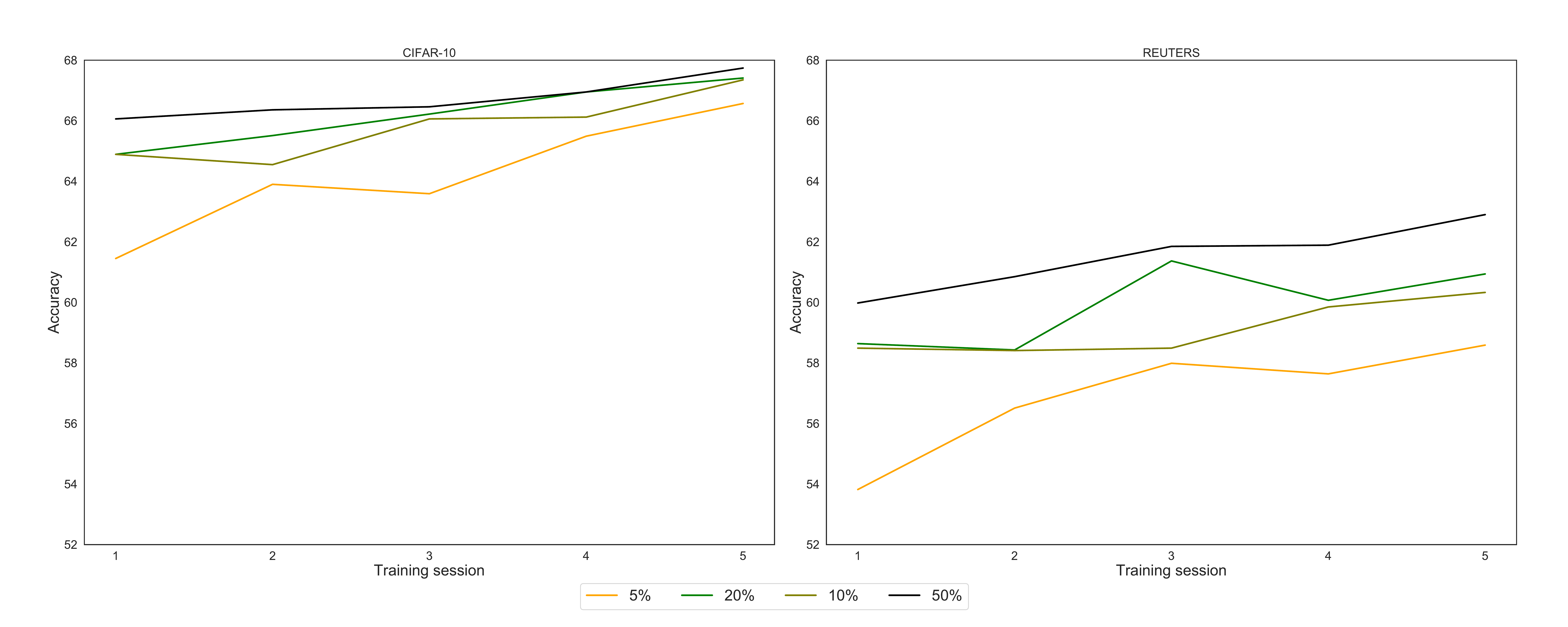}
\caption{Accuracy with respect to the ratio of the number of the samples selected for MAB on CIFAR-10 (left) and REUTERS (right)}
\label{MAB ratio}
\end{figure*}

\begin{figure*}[pt]
\centering
\begin{subfigure}[b]{0.5\textwidth}
  \includegraphics[width =\textwidth]{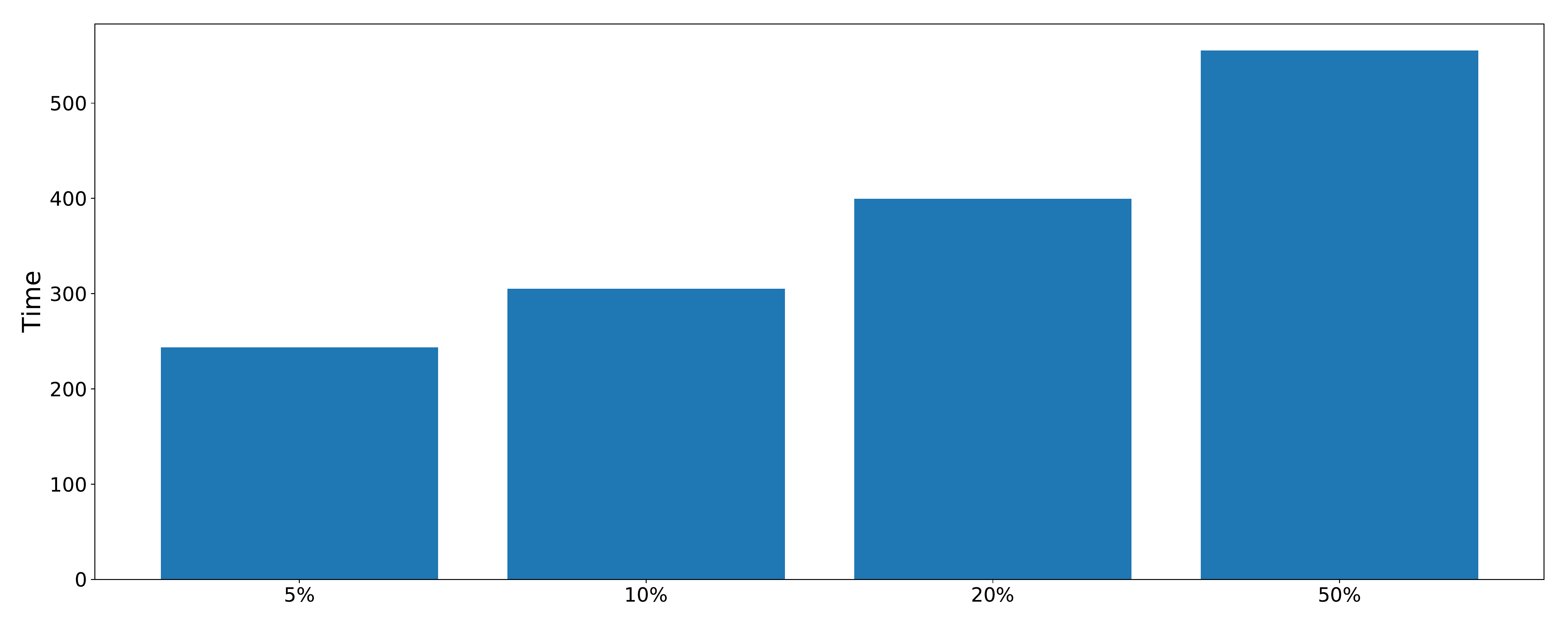}
  \caption{CIFAR-10}
  \label{CIFAR-10 ratio time}
\end{subfigure}%
\begin{subfigure}[b]{0.5\textwidth}
  \includegraphics[width =\textwidth]{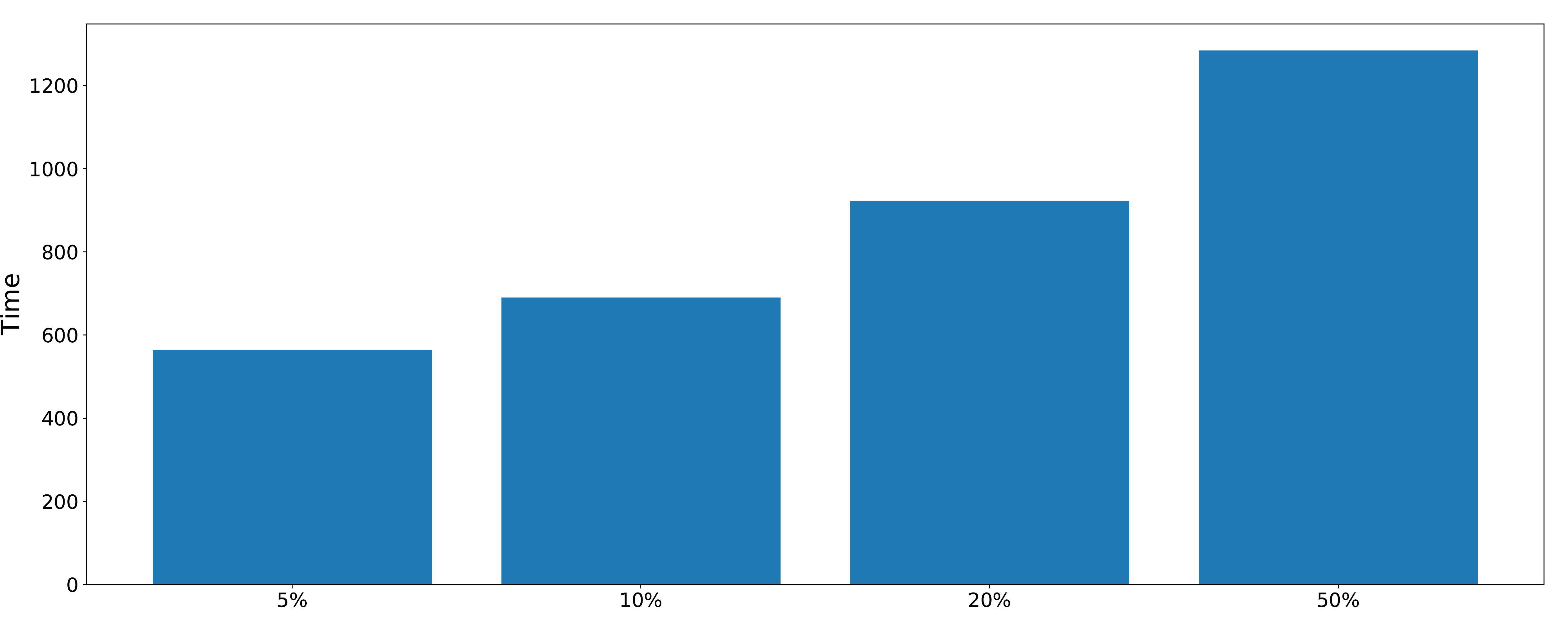}
  \caption{REUTERS}
  \label{REUTERS ratio time}
\end{subfigure}%

 
 
\caption{Average training time in seconds for different sample ratios for MAB}
\label{MAB ratio time}
\end{figure*}

Next in Figure \ref{MAB ratio} we showcase how the size of the selected samples in memory replay based on Algorithm \ref{MAB-based memory replay} affects the test accuracy. The test accuracy increases when the ratio of the sample size over the total training data size increases from 5\% to 50\%. The gap is more pronounced in early sessions. The training time also increases as this sample ratio increases, and we demonstrate in Figure \ref{MAB ratio time}. The running time increases linearly, which is positive.  Even for 50\% it is drastically lower than the corresponding best Union version for these two datasets (the running time of Union (Selfless) for CIFAR-10 is higher than 4,000 seconds and for REUTERS the time of Union (K-FAC + Fine-tune) is more than 13,000 seconds as observed in Table \ref{Best time}). This clearly demonstrates that MAB should be the algorithm of choice. Note that in these two datasets Union outperforms MAB the most in terms of accuracy. 

\begin{figure}[pt]
\begin{subfigure}[b]{0.50\textwidth}
  \includegraphics[width = \textwidth]{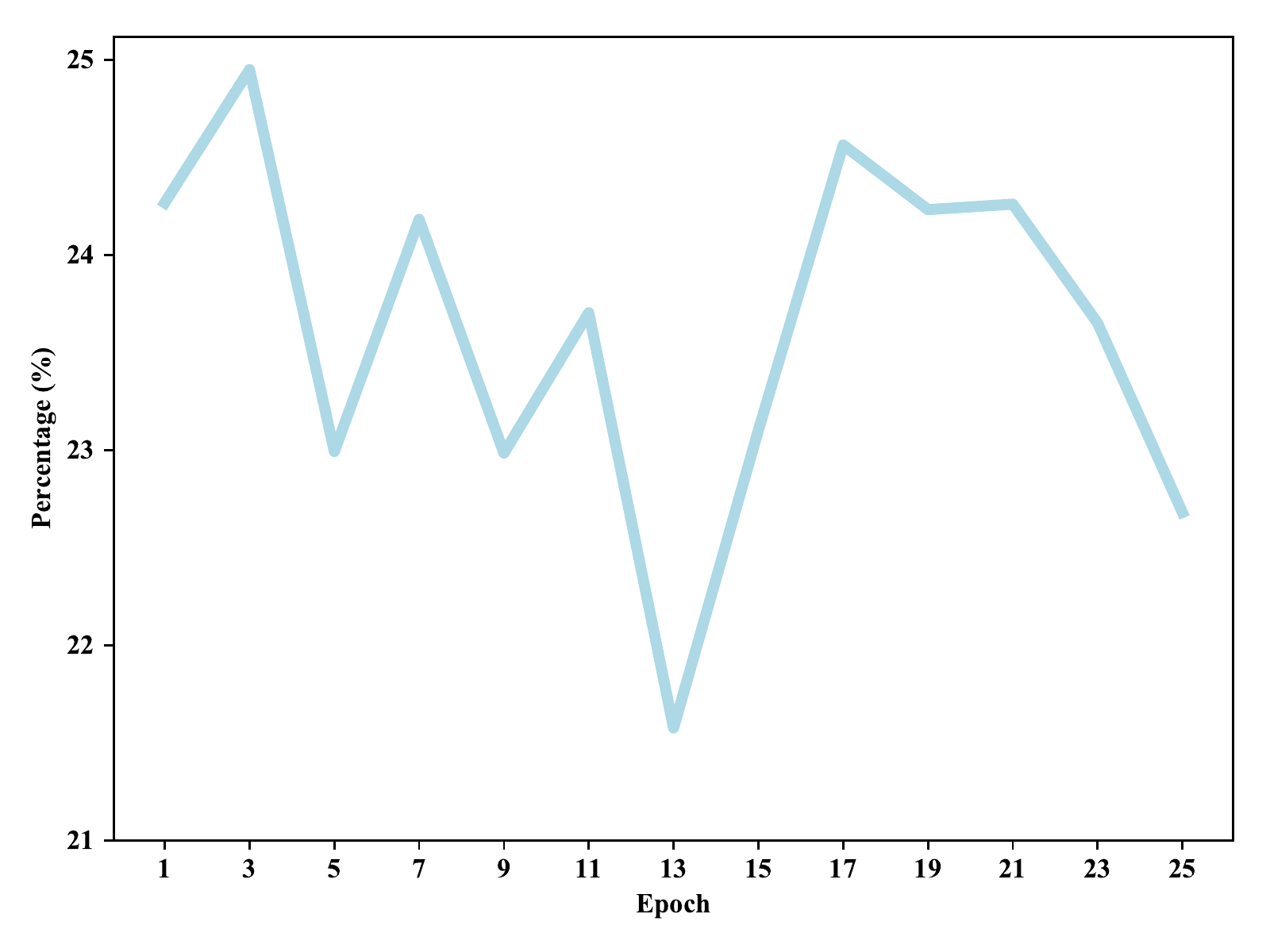}
  \caption{MNIST}
  \label{fig:MNIST MAB weight percentage}
\end{subfigure}
\begin{subfigure}[b]{0.50\textwidth}
  \includegraphics[width =\textwidth]{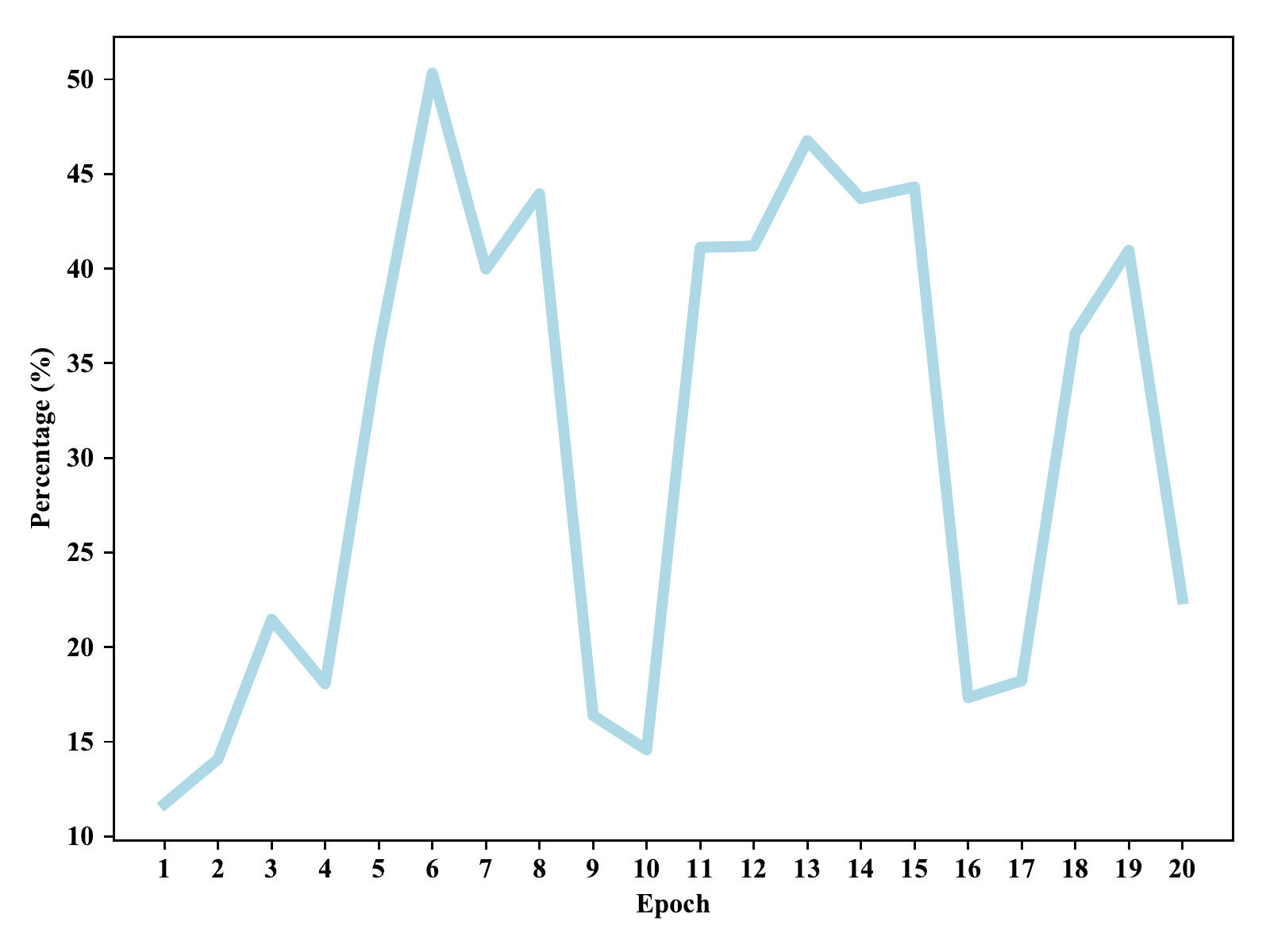}
  \caption{SEA}
  \label{fig:SEA MAB weight percentage}
\end{subfigure}
\begin{subfigure}[b]{0.50\textwidth}
  \includegraphics[width = \textwidth]{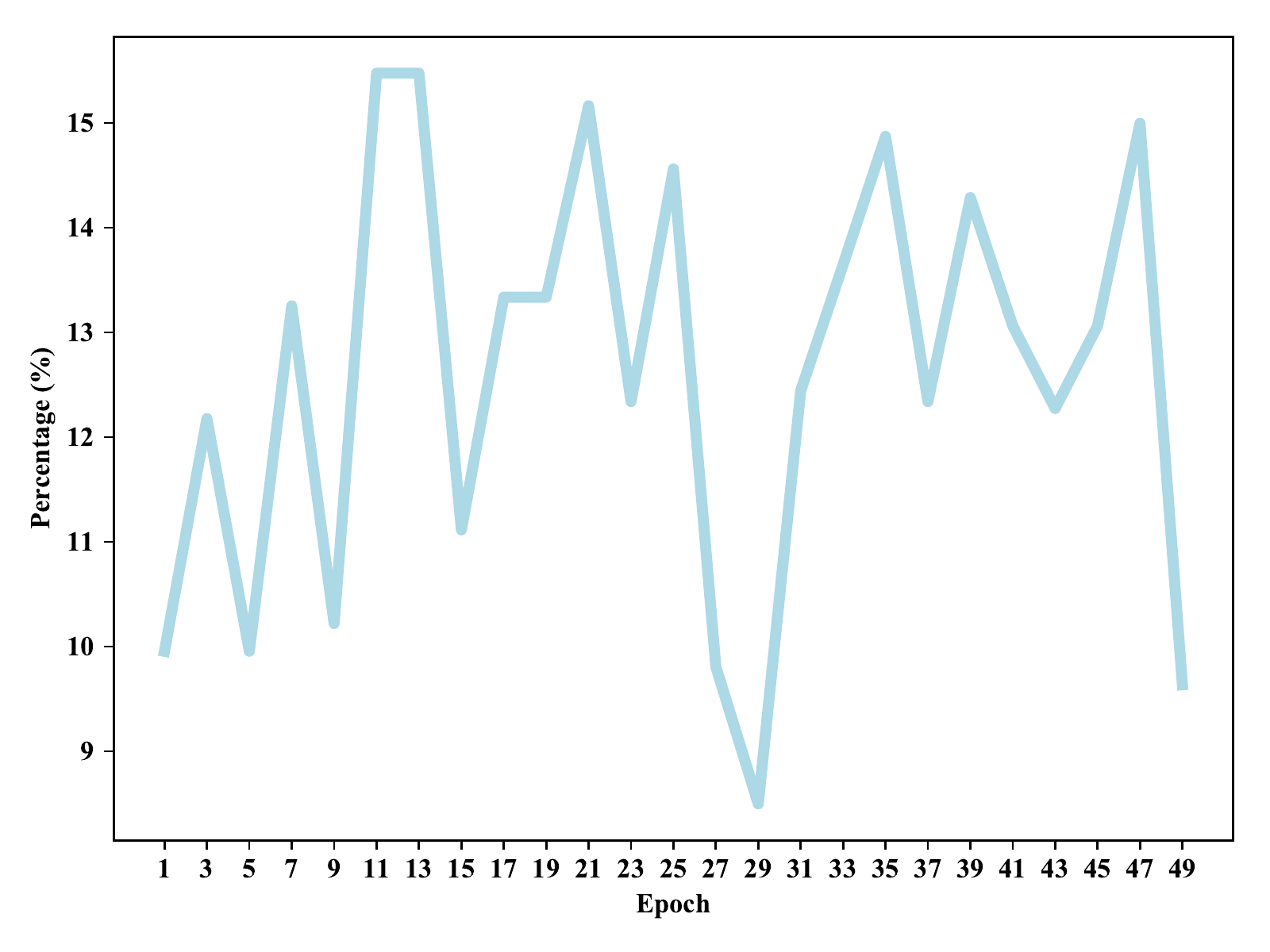}
  \caption{IMDB}
  \label{fig:IMDB MAB weight percentage}
\end{subfigure}
\caption{The average percentage of weights that are optimized in each epoch}
\label{Weight percentage}
\end{figure}

The ratio of the number of weights optimized in every epoch over the total number of weights in a network during a retraining session is illustrated in Figure \ref{Weight percentage}. At most 25\% of weights are optimized in every epoch of the CNN network on the MNIST dataset; at most 50.3\% of the MLP network on the SEA dataset; and at most 16.2\% are optimized in the LSTM network on the IMDB dataset. Similar to dropout, we observe that MAB-based weight sampling may take more epochs before it meets the early stopping criteria compared to the standard epoch-based weight optimization because MAB-based weight sampling can keep searching for a better minimum due to the exploration component while standard retraining soon meets the early stopping criteria.

\begin{figure}[pt]
\begin{subfigure}[b]{0.5\textwidth}
  \includegraphics[width = \textwidth]{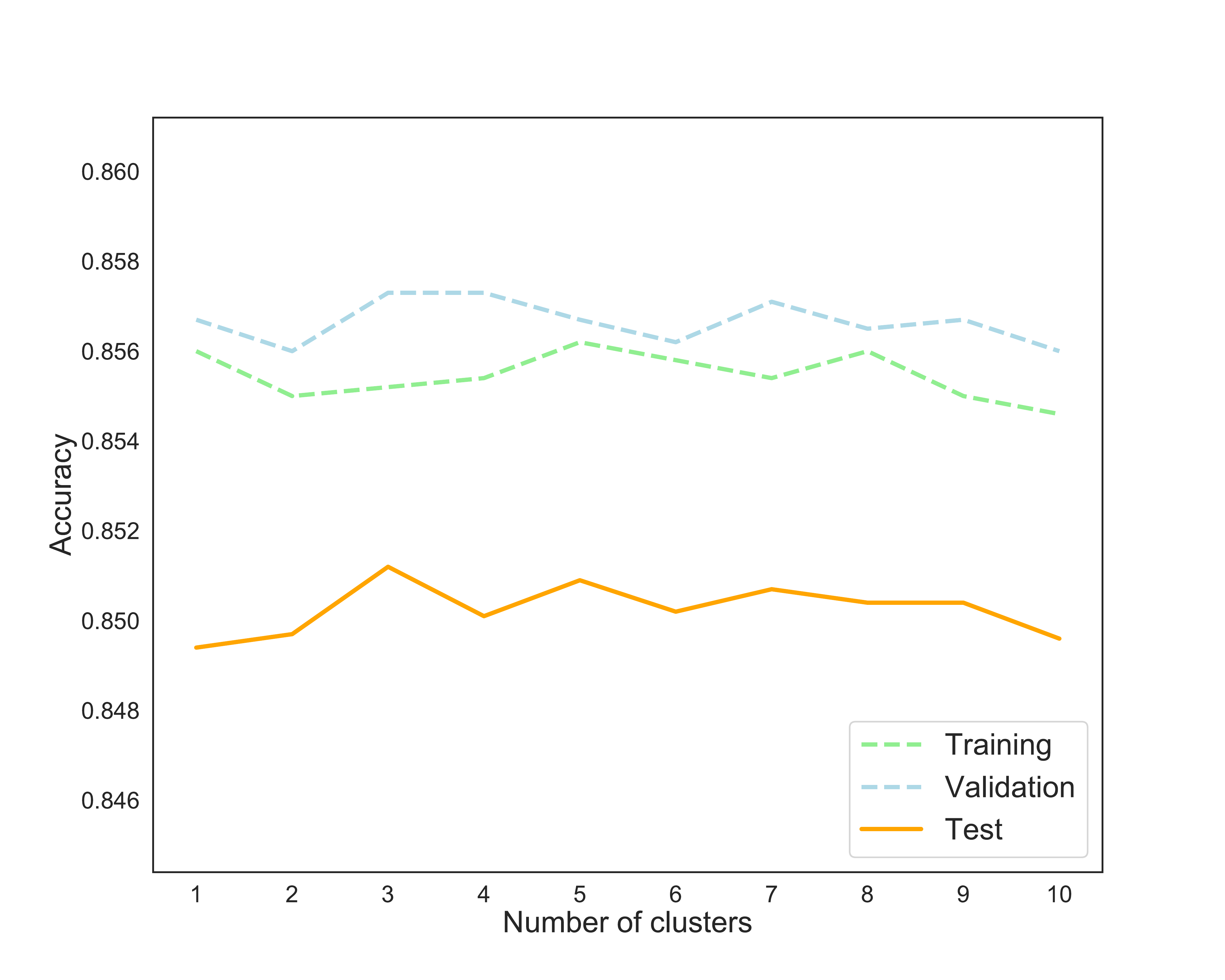}
  \caption{Model accuracy versus the number of clusters}
  \label{fig:Model accuracy and K}
\end{subfigure}%

\begin{subfigure}[b]{0.5\textwidth}
  \includegraphics[width = \textwidth]{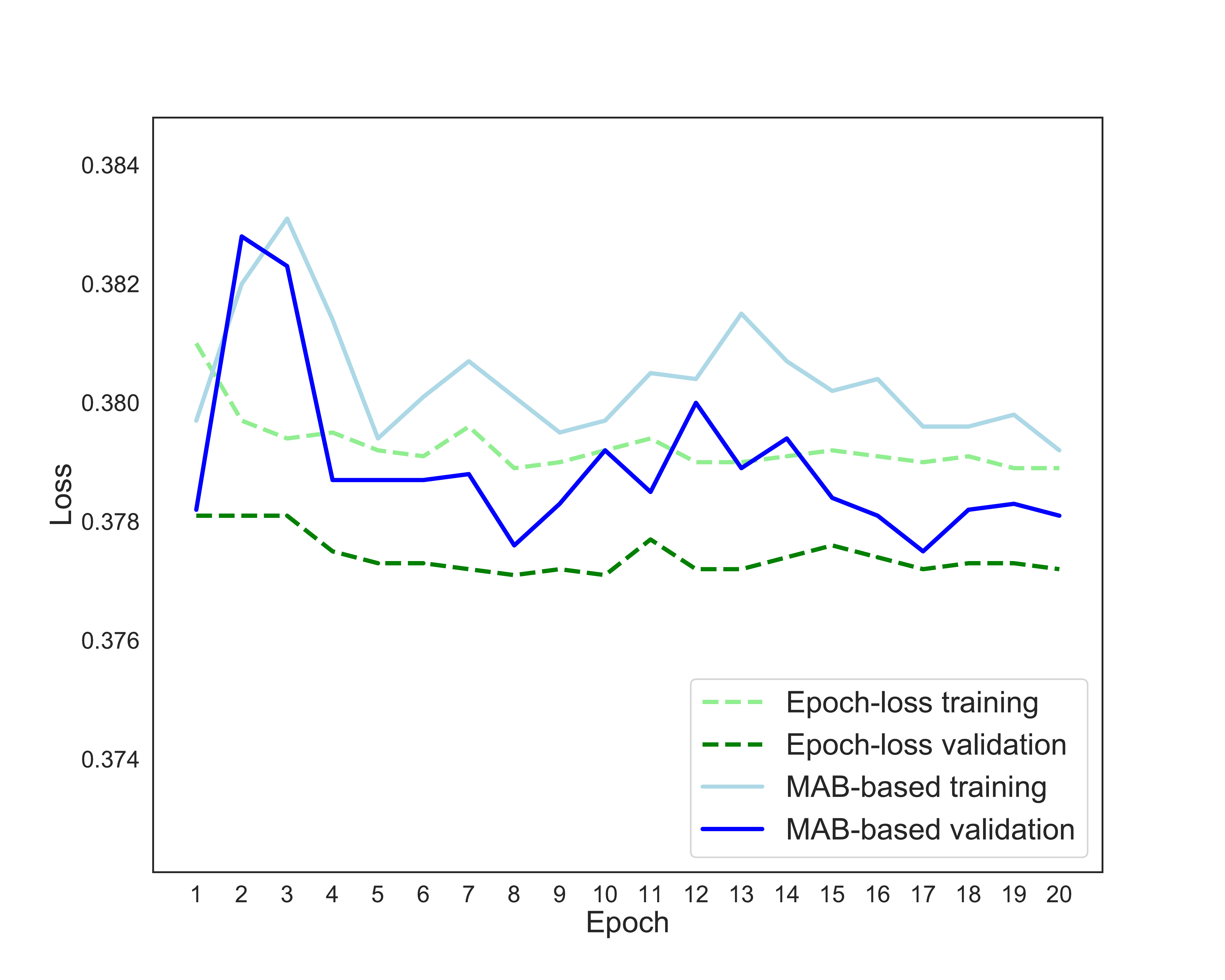}
  \caption{Loss for MAB ($k=3$) and standard epoch-based retraining}
  \label{fig:MAB loss}
\end{subfigure}%

\begin{subfigure}[b]{0.5\textwidth}
  \includegraphics[width =\textwidth]{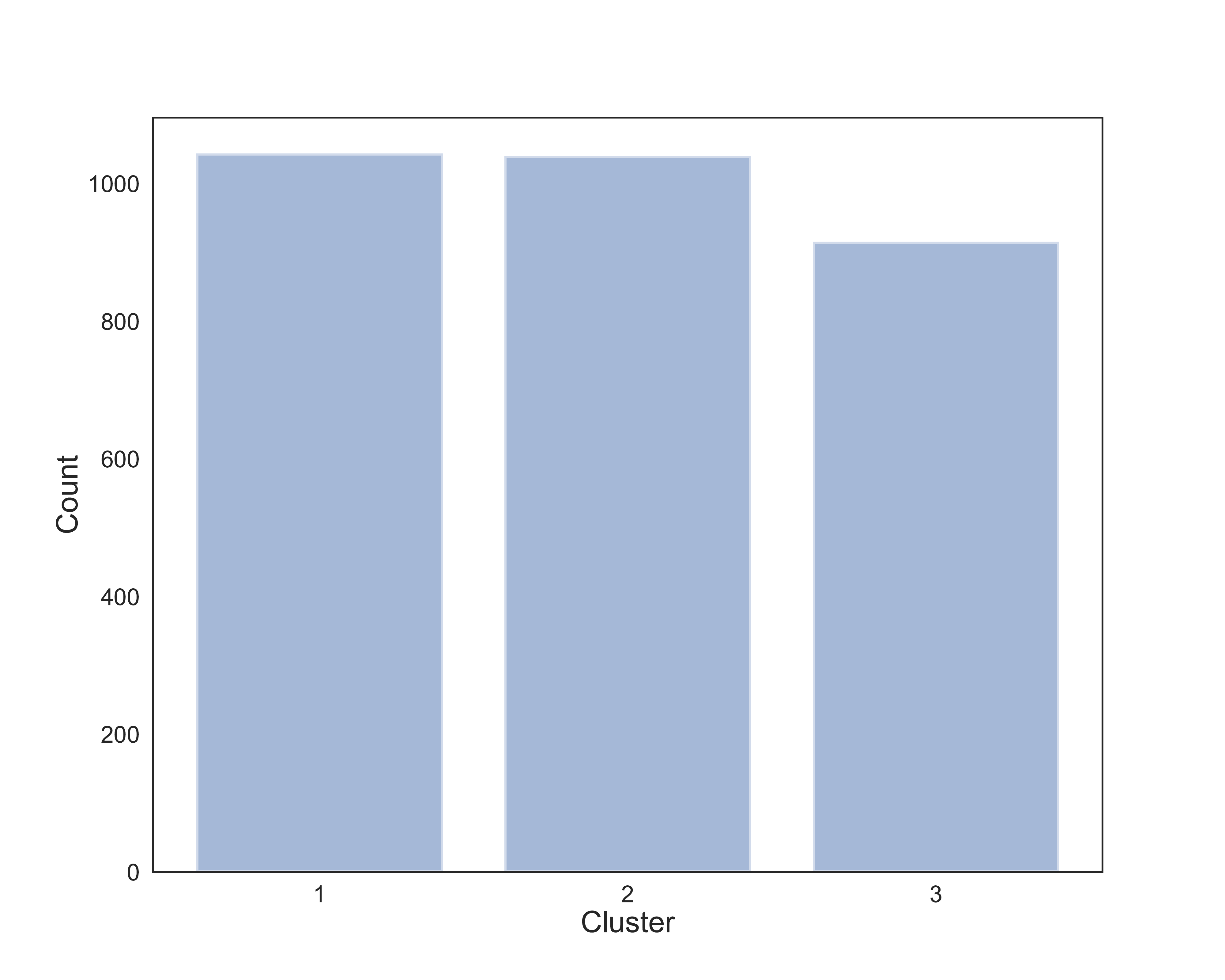}
  \caption{The number of times each cluster is selected ($k=3$)}
  \label{fig:Minibatches and clusters}
\end{subfigure}%
\caption{An MAB-based retraining session of the SEA dataset}
\label{SEA-cluster}
\end{figure}

This is also evident in Figure \ref{fig:MAB loss} that compares the loss of MAB and standard epoch-based training with the same memory replay in the SEA dataset. Loss in MAB is more volatile, which is a further confirmation that MAB explores more. It is also interesting to observe that the training loss of MAB is higher than that of standard epoch-based training. On the other hand, from Tables \ref{Best average accuracy} and \ref{Best relative accuracy} we note that the test performance of MAB is superior, which indicates that MAB generalizes better. The test accuracy is 0.8512 for MAB while it is 0.8497 for standard epoch-based training. This is further explored in Section \ref{4-Model Generalization Datasets}.

We further examine the SEA dataset and MAB as an example to show how the number of clusters in K-Means affects model accuracy as illustrated in Figure \ref{SEA-cluster}. Figure \ref{fig:Model accuracy and K} demonstrates the relationship between the number of clusters in K-Means and model accuracy during training, validation, and test phases. The best validation accuracy is obtained at $k=3$. Thus, our default setting for the number of clusters is 3.
The number of times each cluster of weights is selected during a retraining session is presented in Figure \ref{fig:Minibatches and clusters}. It is clear that each cluster is selected approximately the same number of times. 

We also test the robustness of MAB by utilizing 10 different random seeds on the SEA dataset. The mean accuracy of the ten runs is 0.8511, the standard deviation is 0.0005, the minimal value is 0.8506, and the max value is 0.8522. Very low standard deviation attests to the robustness of the algorithm. 

\subsection{Model Generalization Results}
\label{4-Model Generalization Datasets}
We conduct model generalization experiments similar to those in \cite{simard2003best} by comparing MAB to the standard epoch-based weight optimization utilizing dropout (Dropout) and batch normalization (BN). We compare four training methods, Dropout, BN, Clustering + MAB, and BN + Clustering + MAB, on the same six datasets. (Note that replay buffer has no role here.) In order to demonstrate the model generalization effects of the training methods, we keep the training data unchanged and augment the original test data. For the MNIST and the CIFAR-10 datasets, we use the following popular augmentation factors: image rotations by 45 degrees (clockwise and counterclockwise), image shifting by 20 percent (left and right), and zooming in by 80 to 90 percent. We denote the augmented test datasets by CIFAR-10-A and MNIST-A. In addition, we employ elastic transformation, another popular data augmentation method proposed in \cite{simard2003best}. We denote the transformed test datasets by CIFAR-10-E and MNIST-E. For non-image datasets, e.g., SEA, IMDB, we use the widely used synthetic minority oversampling technique (SMOTE) \cite{chawla2002smote} to add new test examples. In particular, we train using the original training data and test on the combination of the original test data and the augmented test data. The ratio of original test and augmented test data is 50\%. 
For the Clustering + MAB settings, we train using the standard epoch-based weight optimization for at most $x$ epochs and cluster the weights. Then, we train utilizing MAB-based weight optimization to re-optimize weights for the remaining epochs until the training session ends. Dropout and BN are trained using at most $2\cdot x$ epochs. We use the same early stopping criteria as in all of the previous experiments. We set $x$ to be 50 for the MNIST, CIFAR-10, IMDB, and REUTERS datasets and $x$ to be 20 for the SEA and ELEC datasets which are the same values as in previous experiments. 

We show the accuracy results in Table \ref{generation accuracy}. Boldface indicates the highest value in each dataset. The best performance is achieved by combining BN, clustering, and MAB. We find that Clustering+MAB improves model generalization when training a neural network over BN from 0.18\% to 16.0\% with the average improvement being 4.9\%.

\begin{table}[h!]
\centering
\scalebox{0.70}{
\begin{tabular}{|l|r|r|r|r|}\hline
Training Method & CIFAR-10-A & MNIST-A  & CIFAR-10-E & MNIST-E \\ \hline
Dropout                       &67.67                     &98.65                     &43.31                      &88.17                     \\\hline
BN                            &70.54                     &98.88                     &48.74                      &88.20                      \\\hline
Clustering+MAB                &79.16                     &99.09                     &48.92                      &90.29                     \\\hline
BN+Clustering+MAB             &\textbf{80.45}            &\textbf{99.14}            &\textbf{49.23}             &\textbf{90.59}             \\\hline
\hline
Training Method & SEA  & ELEC  & IMDB & REUTERS \\ \hline
Dropout                       &84.09                      &61.98                      &72.07                  &59.66 \\\hline
BN                            &84.39                      &62.14                      &73.24                  &62.92 \\\hline
Clustering+MAB                &84.54                      &72.09                      &76.09                  &65.31 \\\hline
BN+Clustering+MAB             &\textbf{84.88}             &\textbf{73.24}             &\textbf{78.34}         &\textbf{65.61} \\\hline
\end{tabular}
}
\caption{Accuracy (\%) of different training methods for the six datasets}
\label{generation accuracy}
\end{table}

\section{Conclusion}
\label{4-Conclusion}
In this paper, we propose a generic model for continual neural network retraining. Our model integrates neuron importance for encouraging gradient updates for new data, MAB-based memory replay for optimal sampling, and dynamic weight optimization for reducing the number of trainable weights during training and for better generalization. We use various practical data settings to show the robustness of our retraining model in CNN, MLP, and RNN networks. Although we demonstrate the effectiveness of the MAB methodologies for the neural network retraining case, it would be interesting to integrate clustering and MAB-based weight optimization with AutoML. A promising direction to expand our work would be to adjust a trained model when absorbing new features and new classes. 
A convergence property of MAB-based training in the convex and general setting is also of interest. 

\bibliographystyle{named}
\bibliography{ref}

\begin{thebibliography}{}

\bibitem[\protect\citeauthoryear{Aljundi \bgroup \em et al.\egroup
  }{2018a}]{aljundi2018memory}
Rahaf Aljundi, Francesca Babiloni, Mohamed Elhoseiny, Marcus Rohrbach, and
  Tinne Tuytelaars.
\newblock Memory aware synapses: learning what (not) to forget.
\newblock In {\em Proceedings of the European Conference on Computer Vision
  (ECCV)}, pages 139--154, 2018.

\bibitem[\protect\citeauthoryear{Aljundi \bgroup \em et al.\egroup
  }{2018b}]{aljundi2018selfless}
Rahaf Aljundi, Marcus Rohrbach, and Tinne Tuytelaars.
\newblock Selfless sequential learning.
\newblock {\em arXiv preprint arXiv:1806.05421}, 2018.

\bibitem[\protect\citeauthoryear{Audibert and Bubeck}{2010}]{audibert2010best}
Jean-Yves Audibert and S{\'e}bastien Bubeck.
\newblock Best arm identification in multi-armed bandits.
\newblock In {\em Proceedings of the 23rd Annual Conference on Learning Theory
  (COLT)}, 2010.

\bibitem[\protect\citeauthoryear{Auer \bgroup \em et al.\egroup
  }{2002}]{auer2002nonstochastic}
Peter Auer, Nicolo Cesa-Bianchi, Yoav Freund, and Robert~E Schapire.
\newblock The nonstochastic multiarmed bandit problem.
\newblock {\em SIAM Journal on Computing}, 32(1):48--77, 2002.

\bibitem[\protect\citeauthoryear{Auer}{2002}]{auer2002using}
Peter Auer.
\newblock Using confidence bounds for exploitation-exploration trade-offs.
\newblock {\em Journal of Machine Learning Research}, 3(Nov):397--422, 2002.

\bibitem[\protect\citeauthoryear{Beygelzimer \bgroup \em et al.\egroup
  }{2011}]{beygelzimer2011contextual}
Alina Beygelzimer, John Langford, Lihong Li, Lev Reyzin, and Robert Schapire.
\newblock Contextual bandit algorithms with supervised learning guarantees.
\newblock In {\em Proceedings of the Fourteenth International Conference on
  Artificial Intelligence and Statistics (AISTATS)}, pages 19--26, 2011.

\bibitem[\protect\citeauthoryear{Brock \bgroup \em et al.\egroup
  }{2017}]{brock2017freezeout}
Andrew Brock, Theodore Lim, James~M Ritchie, and Nick Weston.
\newblock Freezeout: accelerate training by progressively freezing layers.
\newblock {\em arXiv preprint arXiv:1706.04983}, 2017.

\bibitem[\protect\citeauthoryear{Chawla \bgroup \em et al.\egroup
  }{2002}]{chawla2002smote}
Nitesh~V Chawla, Kevin~W Bowyer, Lawrence~O Hall, and W~Philip Kegelmeyer.
\newblock Smote: synthetic minority over-sampling technique.
\newblock {\em Journal of Artificial Intelligence Research}, 16:321--357, 2002.

\bibitem[\protect\citeauthoryear{Cook and
  Weisberg}{1980}]{cook1980characterizations}
R~Dennis Cook and Sanford Weisberg.
\newblock Characterizations of an empirical influence function for detecting
  influential cases in regression.
\newblock {\em Technometrics}, 22(4):495--508, 1980.

\bibitem[\protect\citeauthoryear{French}{1999}]{french1999catastrophic}
Robert~M French.
\newblock Catastrophic forgetting in connectionist networks.
\newblock {\em Trends in Cognitive Sciences}, 3(4):128--135, 1999.

\bibitem[\protect\citeauthoryear{Ge \bgroup \em et al.\egroup
  }{2015}]{ge2015escaping}
Rong Ge, Furong Huang, Chi Jin, and Yang Yuan.
\newblock Escaping from saddle points - online stochastic gradient for tensor
  decomposition.
\newblock In {\em Conference on Learning Theory (COLT)}, pages 797--842, 2015.

\bibitem[\protect\citeauthoryear{Graves \bgroup \em et al.\egroup
  }{2017}]{graves2017automated}
Alex Graves, Marc~G Bellemare, Jacob Menick, Remi Munos, and Koray Kavukcuoglu.
\newblock Automated curriculum learning for neural networks.
\newblock In {\em Proceedings of the 34th International Conference on Machine
  Learning (ICML)}, pages 1311--1320. JMLR. org, 2017.

\bibitem[\protect\citeauthoryear{Han \bgroup \em et al.\egroup
  }{2016}]{han2015deep_compression}
Song Han, Huizi Mao, and William~J Dally.
\newblock Deep compression: compressing deep neural networks with pruning,
  trained quantization and huffman coding.
\newblock In {\em International Conference on Learning Representations (ICLR)},
  2016.

\bibitem[\protect\citeauthoryear{Kaufmann \bgroup \em et al.\egroup
  }{2012}]{kaufmann2012thompson}
Emilie Kaufmann, Nathaniel Korda, and R{\'e}mi Munos.
\newblock Thompson sampling: an asymptotically optimal finite-time analysis.
\newblock In {\em International Conference on Algorithmic Learning Theory},
  pages 199--213, 2012.

\bibitem[\protect\citeauthoryear{Kemker \bgroup \em et al.\egroup
  }{2018}]{kemker2018measuring}
Ronald Kemker, Marc McClure, Angelina Abitino, Tyler~L Hayes, and Christopher
  Kanan.
\newblock Measuring catastrophic forgetting in neural networks.
\newblock In {\em Thirty-second AAAI Conference on Artificial Intelligence},
  2018.

\bibitem[\protect\citeauthoryear{Kilinc and Uysal}{2017}]{kilinc2017auto}
Ozsel Kilinc and Ismail Uysal.
\newblock Auto-clustering output layer: automatic learning of latent
  annotations in neural networks.
\newblock {\em arXiv preprint arXiv:1702.08648}, 2017.

\bibitem[\protect\citeauthoryear{Kirkpatrick \bgroup \em et al.\egroup
  }{2017}]{kirkpatrick2017overcoming}
James Kirkpatrick, Razvan Pascanu, Neil Rabinowitz, Joel Veness, Guillaume
  Desjardins, Andrei~A Rusu, Kieran Milan, John Quan, Tiago Ramalho, Agnieszka
  Grabska-Barwinska, et~al.
\newblock Overcoming catastrophic forgetting in neural networks.
\newblock {\em Proceedings of the National Academy of Sciences},
  114(13):3521--3526, 2017.

\bibitem[\protect\citeauthoryear{Koh and Liang}{2017}]{koh2017understanding}
Pang~Wei Koh and Percy Liang.
\newblock Understanding black-box predictions via influence functions.
\newblock In {\em Proceedings of the 34th International Conference on Machine
  Learning (ICML)}, pages 1885--1894, 2017.

\bibitem[\protect\citeauthoryear{Krizhevsky \bgroup \em et al.\egroup
  }{2009}]{krizhevsky2009learning}
Alex Krizhevsky, Geoffrey Hinton, et~al.
\newblock Learning multiple layers of features from tiny images.
\newblock Technical report, 2009.

\bibitem[\protect\citeauthoryear{Li and Hoiem}{2017}]{li2017learning}
Zhizhong Li and Derek Hoiem.
\newblock Learning without forgetting.
\newblock {\em IEEE Transactions on Pattern Analysis and Machine Intelligence},
  40(12):2935--2947, 2017.

\bibitem[\protect\citeauthoryear{Lopez-Paz and
  Ranzato}{2017}]{lopez2017gradient}
David Lopez-Paz and {Marc'Aurelio} Ranzato.
\newblock Gradient episodic memory for continual learning.
\newblock In {\em Advances in Neural Information Processing Systems (NeurIPS)},
  pages 6467--6476, 2017.

\bibitem[\protect\citeauthoryear{Mehta \bgroup \em et al.\egroup
  }{2019}]{Mehta2018}
Sanket~Vaibhav Mehta, Bhargavi Paranjape, and Sumeet Singh.
\newblock Evaluating influence functions for memory replay in continual
  learning.
\newblock In {\em International Conference on Machine Learning (ICML)}, 2019.

\bibitem[\protect\citeauthoryear{Qin \bgroup \em et al.\egroup
  }{2017}]{qin2017improving}
Chao Qin, Diego Klabjan, and Daniel Russo.
\newblock Improving the expected improvement algorithm.
\newblock In {\em Advances in Neural Information Processing Systems (NeurIPS)},
  pages 5381--5391, 2017.

\bibitem[\protect\citeauthoryear{Rusu \bgroup \em et al.\egroup
  }{2016}]{Rusu2016}
Andrei~A. Rusu, Neil~C. Rabinowitz, Guillaume Desjardins, Hubert Soyer, James
  Kirkpatrick, Koray Kavukcuoglu, Razvan Pascanu, and Raia Hadsell.
\newblock Progressive neural networks.
\newblock {\em CoRR}, 2016.

\bibitem[\protect\citeauthoryear{Saito and
  Nakano}{2007}]{saito2007bidirectional}
Kazumi Saito and Ryohei Nakano.
\newblock Bidirectional clustering of weights for neural networks with common
  weights.
\newblock {\em Systems and Computers in Japan}, 38(10):46--57, 2007.

\bibitem[\protect\citeauthoryear{Simard \bgroup \em et al.\egroup
  }{2003}]{simard2003best}
Patrice~Y Simard, David Steinkraus, John~C Platt, et~al.
\newblock Best practices for convolutional neural networks applied to visual
  document analysis.
\newblock In {\em Proceedings of the Seventh International Conference on
  Document Analysis and Recognition (ICDAR)}, volume~3, 2003.

\bibitem[\protect\citeauthoryear{Swaroop \bgroup \em et al.\egroup
  }{2019}]{swaroop2019improving}
Siddharth Swaroop, Cuong~V. Nguyen, Thang~D. Bui, and Richard~E. Turner.
\newblock Improving and understanding variational continual learning.
\newblock {\em arXiv preprint arXiv:1905.02099}, 2019.

\bibitem[\protect\citeauthoryear{Vitter}{1985}]{vitter1985random}
Jeffrey~S Vitter.
\newblock Random sampling with a reservoir.
\newblock {\em ACM Transactions on Mathematical Software}, 11(1):37--57, 1985.

\bibitem[\protect\citeauthoryear{Wang \bgroup \em et al.\egroup
  }{2017}]{wang2017accelerating}
Linnan Wang, Yi~Yang, Renqiang Min, and Srimat Chakradhar.
\newblock Accelerating deep neural network training with inconsistent
  stochastic gradient descent.
\newblock {\em Neural Networks}, 93:219--229, 2017.

\bibitem[\protect\citeauthoryear{Wu \bgroup \em et al.\egroup
  }{2018}]{wu2018deep}
Junru Wu, Yue Wang, Zhenyu Wu, Zhangyang Wang, Ashok Veeraraghavan, and Yingyan
  Lin.
\newblock Deep $ k $-means: re-training and parameter sharing with harder
  cluster assignments for compressing deep convolutions.
\newblock {\em arXiv preprint arXiv:1806.09228}, 2018.

\bibitem[\protect\citeauthoryear{Yu \bgroup \em et al.\egroup
  }{2019}]{YuSlimmable}
Jiahui Yu, Linjie Yang, Ning Xu, Jianchao Yang, and Thomas Huang.
\newblock Slimmable neural networks.
\newblock {\em International Conference on Learning Representations (ICLR)},
  2019.

\end{thebibliography}

\end{document}